\documentclass{article}

\usepackage{microtype}
\usepackage{graphicx}
\usepackage{subcaption}
\usepackage{booktabs} 

\usepackage{hyperref}
\usepackage{comment}

 \usepackage[preprint]{icml2026}

\usepackage{amsmath}
\usepackage{amssymb}
\usepackage{mathtools}
\usepackage{amsthm}

\usepackage[capitalize,noabbrev]{cleveref}

\theoremstyle{plain}

\theoremstyle{definition}

\theoremstyle{remark}

\usepackage[textsize=tiny]{todonotes}

\icmltitlerunning{Bus-Conditioned Zero-Shot Trajectory Generation via Task Arithmetic}

\begin{document}

\twocolumn[
  \icmltitle{Bus-Conditioned Zero-Shot Trajectory Generation via Task Arithmetic}



  \icmlsetsymbol{equal}{*}

  \begin{icmlauthorlist}
    \icmlauthor{Shuai Liu}{yyy}
    \icmlauthor{Ning Cao}{yyy}
    \icmlauthor{Yile Chen}{yyy}
    \icmlauthor{Yue Jiang}{yyy}
    \icmlauthor{Gao Cong}{yyy}
  \end{icmlauthorlist}

  \icmlaffiliation{yyy}{College of Computing and Data Science, Nanyang Technological University, Singapore}

  \icmlkeywords{Machine Learning, ICML}

  \icmlcorrespondingauthor{Shuai Liu}{Shuai004@e.ntu.edu.sg}

  \vskip 0.3in
]



\printAffiliationsAndNotice{}  
\begin{abstract}
    Mobility trajectory data provide essential support for smart city applications.
    However, such data are often difficult to obtain. 
    Meanwhile, most existing trajectory generation methods implicitly assume that at least a subset of real mobility data from  target city is available, which limits their applicability in data-inaccessible scenarios. 
    In this work, we propose a new problem setting, called bus-conditioned zero-shot trajectory generation, where no mobility trajectories from a target city are accessible.
    The generation process relies solely on source city mobility data and publicly available bus timetables from both cities.
Under this setting, we propose MobTA, the first approach to introduce task arithmetic into trajectory generation. 
MobTA  models the parameter shift from bus-timetable-based trajectory generation to mobility trajectory generation in source city, and applies this shift to target city through arithmetic operations on task vectors.
This enables trajectory generation that reflects target-city mobility patterns without requiring any real mobility data from it. 
Furthermore, we  theoretically analyze MobTA's stability  across base and instruction-tuned LLMs.
Extensive experiments show that MobTA significantly outperforms existing methods, and achieves performance close to models finetuned using target city mobility trajectories.
\end{abstract}

\section{Introduction}
Human mobility trajectories characterize the spatiotemporal behavior of urban residents~\cite{liu2024nextlocllm,wei2025transfertraj}, supporting various smart city applications such as urban planning~\cite{zhou2024large}, traffic management~\cite{lu2019trajectory}, and disease control~\cite{pollington2025evaluation}.
In practice, however, real-world mobility trajectories are often 
difficult to access.
First, such data are typically collected by  ride-hailing platforms or telecommunication operators, and are rarely publicly accessible due to commercial restrictions. 
 Second, mobility trajectories contain sensitive personal information, and their  sharing are subject to strict legal and ethical constraints.
These challenges motivate trajectory generation, which aims to model the  spatiotemporal patterns of human mobility based on available trajectory data, and to generate synthetic trajectories that 
closely match real-world mobility~\cite{chen2025trajectory}.

\begin{figure}
    \centering
    \includegraphics[scale=0.35]{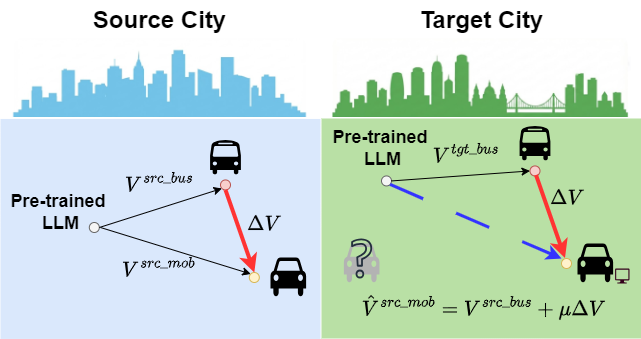}
    \caption{
        Conceptual overview of  MobTA.
        It uses task arithmetic to model the parameter shift from bus-timetable-based trajectory generation to  mobility trajectory generation in the source city, and applies this shift to the target city, where only bus timetables are accessible, to enable mobility trajectory generation for target city.
    }
    \label{fig:brief essense}
    \vspace{-15pt}
\end{figure}

Although existing trajectory generation models have made notable progress, most of them rely on an implicit assumption: at least a subset of real mobility trajectories from target city is accessible during training. 
However, satisfying this prerequisite is often impractical.
Beyond the aforementioned commercial restrictions and privacy concerns, many cities are unable to collect and maintain mobility trajectory data due to the substantial costs of sensor deployment and the long-term burden of data management~\cite{kong2023mobility}. 
As a result, existing methods are difficult to apply directly in such data-inaccessible cities.
In contrast, bus timetable data offer clear advantages in terms of accessibility. 
As non-sensitive public data, bus timetables are widely available across mapping platforms like Google Map, Amap, and Naver Map.
Moreover, Moovit~\footnote{\url{https://moovitapp.com/}} 
collects bus timetable data across multiple countries.
Each bus timetable consists of an ordered sequence of bus stops and their scheduled arrival times, which can be regarded as a structured yet coarse-grained form of trajectory data.
Nevertheless, bus timetables alone cannot effectively substitute mobility trajectories in trajectory generation. 
\cite{zhang2018different} has shown that while the two types of trajectories exhibit certain correlations, gaps remain in terms of spatial coverage and distribution. 
Constrained by fixed routes, bus timetables fail to capture the flexible travel behaviors that occur beyond bus routes. 
This 
motivates our core
research question:
In the absence of 
real mobility trajectories from target city, how can we leverage  mobility trajectories from other cities, together with publicly available bus timetables, to generate synthetic trajectories that 
reflect the  mobility patterns of target city?
We 
define this challenging yet practical problem setting as \textbf{bus-conditioned zero-shot trajectory generation}.

To address this problem, our core intuition is that bus routes are designed to cover the main urban travel demands, aiming to improve overall mobility efficiency and alleviate traffic congestion~\cite{vuchic2017urban}. 
Thus, although bus timetables cannot fully capture urban mobility behaviors, they represent a mobility 'backbone' in terms of spatial layout and temporal regularity, reflecting  activity hotspots and dominant travel directions.
Mobility trajectories, in turn, provide more flexible and fine-grained mobility patterns that complement this backbone.
We further posit that the discrepancy between bus timetables and mobility trajectories is a structured deviation primarily arising from first-and-last-mile travel and personalized short-range movements~\cite{lu2024first}.
As such behaviors commonly appear  with similar structural patterns~\cite{gonzalez2008understanding}, this discrepancy exhibits a certain degree of transferability across cities, highlighting the need for a mechanism capable of explicitly modeling and transferring such structured deviations.

Motivated by this intuition, we draw inspiration from task arithmetic, which has recently emerged as a promising paradigm for model editing~\cite{ilharco2022editing,li2025task}.
Its core concept is  task vector, which is defined as the parameter difference between a fine-tuned model and its original pre-trained model. 
Prior studies~\cite{bhardwaj2024language,zhang2024knowledge} have shown that linear arithmetic  operations on task vectors exhibits great effectiveness in out-of-domain generalization and has achieved success across a variety of tasks~\cite{parovic2024investigating,braga2025investigating}. 
These findings suggest that, even when the target task or target domain is not explicitly optimized during training, a model may still be able to have target task-relevant capabilities through appropriate combinations of task vectors.

Building upon the above intuition and the task arithmetic paradigm,  we propose \textbf{Mob}ility \textbf{T}ask \textbf{A}rithmetic (MobTA), a novel method to address bus-conditioned zero-shot trajectory generation. 
It constructs three task vectors corresponding to mobility trajectory generation task in source city, and bus-timetable-based trajectory generation task in both source and target cities. 
As shown in Fig.~\ref{fig:brief essense}, MobTA  uses task arithmetic to model the parameter shift from bus-timetable-based trajectory generation to  mobility trajectory generation in the source city.
It then applies this shift to the target city’s bus-timetable-based trajectory generation task vector. 
This mechanism enables trajectory generation that approximates  target city’s mobility patterns, without requiring access to its real mobility data.
Furthermore, we provide a theoretical analysis of MobTA's stability, explaining why task vectors obtained by fine-tuning either base models or instruction-tuned models lead to similar trajectory generation behaviors when combined through task arithmetic.
Experimental results show that MobTA consistently outperforms existing baselines, and achieves performance close to that obtained by fine-tuning models with target city mobility trajectories.
In summary, our contributions are:

\begin{itemize}
    \item We introduce bus-conditioned zero-shot trajectory generation, a challenging yet practical  problem setting where  target city mobility trajectories are entirely inaccessible, and trajectory generation relies solely on mobility trajectories from other cities, and publicly accessible bus timetables.
This offers a new perspective for trajectory generation in data-inaccessible cities.

\item We propose MobTA, the first model to apply task arithmetic to trajectory generation.
It captures the parameter shift from bus-timetable-based to mobility trajectory generation in the source city, and applies this shift to the target city for  mobility trajectory generation. We further theoretically analyze MobTA's stability across base and instruction-tuned LLMs.

\item Extensive experiments demonstrate that MobTA outperforms existing baselines, and achieves trajectory generation performance close to that obtained by fine-tuning with real mobility trajectories from  target city.

\end{itemize}

\section{Related Work}
\subsection{Trajectory Generation}
Trajectory generation aims to synthesize  mobility data by learning spatiotemporal patterns from real trajectories. 
Early methods typically model human mobility using Markov processes~\cite{mathew2012predicting,baratchi2014hierarchical}, gravity model~\cite{pappalardo2015returners}, or social interaction graphs~\cite{borrel2008simps}. 
While such methods provide interpretable formulations, they struggle to capture the complex patterns and heterogeneous behaviors in real-world mobility data.
In recent years, learning-based approaches have become the dominant paradigm. 
Methods based on generative adversarial networks (GANs) leverage adversarial training to align generated distributions with real data~\cite{feng2020learning,rao2020lstm,cao2021generating}, while variational autoencoders (VAEs) model trajectories through structured latent representations~\cite{huang2019variational,ding2019multi}. 
Recently, diffusion models have been introduced to trajectory generation.
They produce high-fidelity trajectories through iterative denoising~\cite{zhu2023difftraj,chu2024simulating,zhu2024controltraj}.
For a detailed review of trajectory generation models, please refer to Appendix~\ref{sec:detailed traj}.
 Despite their effectiveness, most existing methods implicitly assume that at least a subset of target city mobility trajectories  is accessible.
In practice, however, real mobility trajectory data are often difficult to obtain due to privacy concerns or commercial restrictions. 
As a result, existing approaches cannot be directly applied to such data-inaccessible cities.


\subsection{Task Arithmetic}
Task arithmetic is a parameter-efficient paradigm for model editing, enabling the injection of new abilities or the mitigation of undesirable behaviors without costly retraining~\cite{yadav2023ties,bhardwaj2024language}.
Its core concept is the task vector, which is defined as the parameter difference between a fine-tuned model and its pre-trained counterpart~\cite{ilharco2022editing,zhang2023composing}.
More importantly, task arithmetic exhibits remarkable potential for out-of-distribution (OOD) generalization, enabling models to adapt to novel tasks without additional training~\cite{chronopoulou2024language,parovic2024investigating} or narrowing distribution gaps between synthetic and real data~\cite{su2024task,braga2025investigating}.
Theoretically, \cite{li2025task} validate that linear compositions of task vectors can provably guide models toward such generalization, providing a rigorous foundation for this paradigm. 
A detailed review of task arithmetic is provided in Appendix.~\ref{sec:detail ta}.
 
\section{Preliminary}
\subsection{Problem Formulation}

\textit{Definition 1 (\textbf{Mobility Trajectory})}
A mobility trajectory is an ordered sequence of location-time pairs, denoted as $\tau^{mob} = \{(l_1, t_1), (l_2, t_2), \dots, (l_n, t_n)\}$, where $l_i$ is the spatial location  at time $t_i$.
Given a city $c$, its mobility trajectory data is denoted as
$\mathcal{D}^{mob}_c = \{\tau_1^{mob}, \tau_2^{mob}, \dots, \tau_N^{mob}\}$.

\textit{Definition 2 (\textbf{Bus Timetable})}
A bus timetable  is defined  as an ordered sequence of station-arrival time pairs
$\tau^{bus} = \{(s_1, \tilde{t}_1), (s_2, \tilde{t}_2), \dots, (s_m, \tilde{t}_m)\}$,
where $s_i$ denotes the $i$-th bus station on the route and $\tilde{t}_i$ denotes the scheduled arrival time at station $s_i$. 
Given a city $c$, its bus timetable data is denoted as
$\mathcal{D}^{bus}_c = \{\tau_1^{bus}, \tau_2^{bus}, \dots, \tau_N^{bus}\}$.

\textit{Definition 3 (\textbf{Bus Station Sequence})}
Given a bus timetable $\tau^{bus}$, its corresponding bus station sequence is defined as the ordered list of bus stations
$r^{bus} = (s_1, s_2, \dots, s_m)$.

\textit{Definition 4 (\textbf{Bus-Conditioned Zero-Shot Trajectory Generation})}
Bus-conditioned zero-shot trajectory generation considers a problem setting where no mobility trajectories from target city $c_{tgt}$ are accessible during training.
Available data consist of source city data $(\mathcal{D}^{mob}_{c_{src}},\mathcal{D}^{bus}_{c_{src}})$ and target city bus timetables $\mathcal{D}^{bus}_{c_{tgt}}$. 
The objective is to obtain a trajectory generation model $\mathcal{G}$ that produces  synthetic trajectories $\hat{\mathcal{D}}^{mob}_{c_{tgt}}$ that reflect the mobility patterns of  $c_{tgt}$.
Formally,
\begin{equation}
    \hat{\mathcal{D}}^{mob}_{c_{tgt}}=\mathcal{G}(\mathcal{D}^{mob}_{c_{src}},\mathcal{D}^{bus}_{c_{src}},\mathcal{D}^{bus}_{c_{tgt}})
\end{equation}

\subsection{Empirical Observations on Bus Timetables and Mobility Trajectories}
\label{sec:emp}

We conduct an empirical analysis of bus timetables and their relationship with  mobility trajectories. 
We focus on two  aspects: (1) the long-term spatial stability of bus station sequences, and (2) the spatiotemporal correlations between bus timetables and mobility trajectories.
These empirical findings provide practical motivation for explicitly modeling the parameter shift from bus-timetable-based trajectory generation to general mobility trajectory generation in MobTA.

\subsubsection{Long-Term Spatial Stability of Bus Station Sequences}

To examine the long-term spatial stability of bus station sequences, we conduct a comparative analysis using Shanghai bus data in 2015 and 2025.
As the 2015 data lacks scheduled arrival time, our analysis focuses solely on bus station sequences.
We evaluate the similarity of these sequences using multiple sequence similarity metrics. 

\begin{table}[ht]
  \centering
  \caption{Long-term Spatial Stability of Shanghai bus station sequences in 2015 and 2025.}
  \label{sh bus 2015}
  \vskip 0.1in
  \begin{small}
    \setlength{\tabcolsep}{3.2pt} 
    \begin{tabular}{lccccc}
      \toprule
      Category & LCS($\uparrow$) & Jac($\uparrow$) & EDR($\downarrow$) & DTW($\downarrow$) & Dice($\uparrow$) \\
      \midrule
    Entire City     & 0.8109 & 0.6975 & 0.2522 & 0.1355 & 0.7937 \\
      Within Outer Ring  & 0.8546 & 0.7459 & 0.2011 & 0.1062 & 0.8383 \\
      \bottomrule
    \end{tabular}
  \end{small}
  \vskip -0.1in
\end{table}

\begin{figure*}
  \centering
  \includegraphics[scale=0.23]{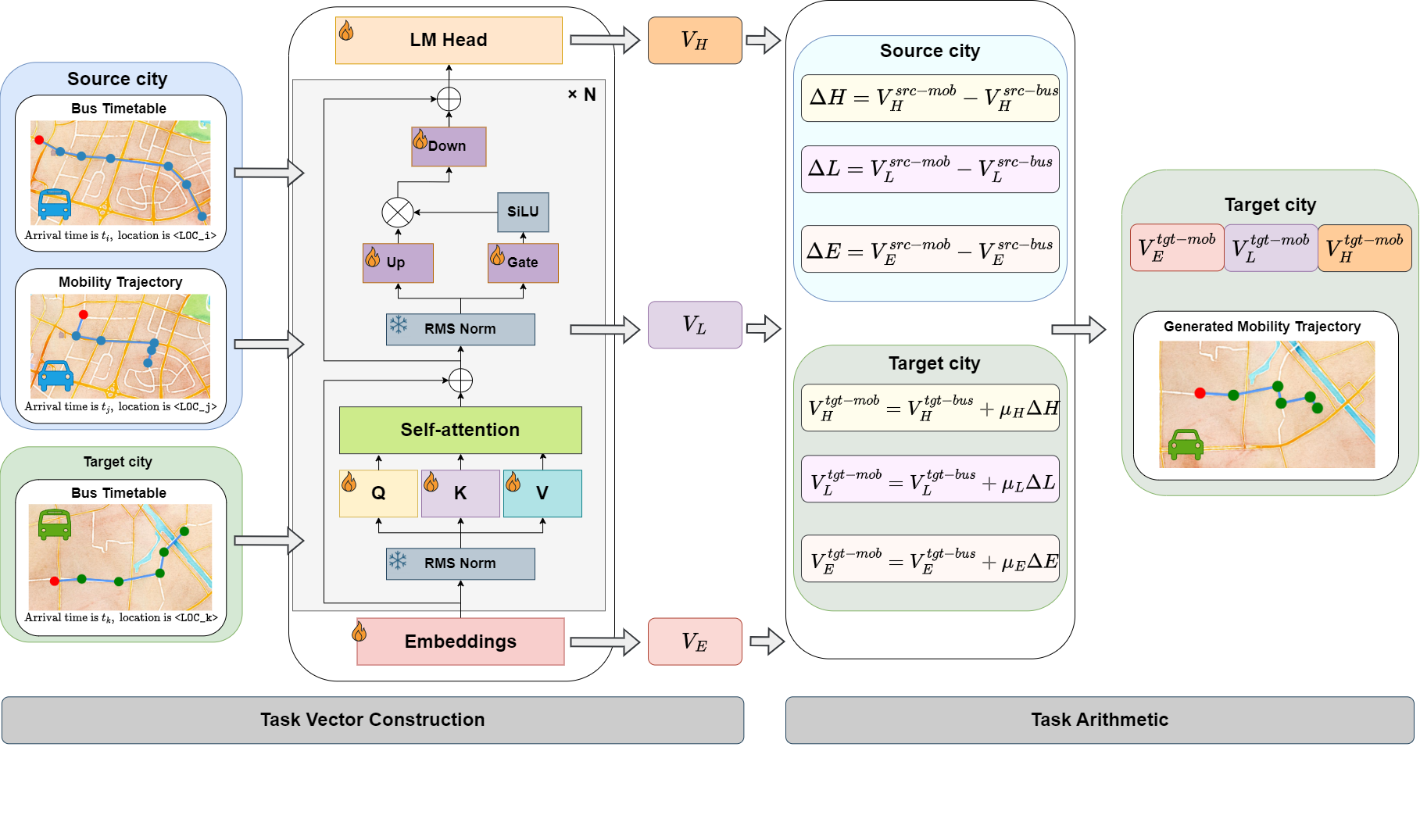}
  \vspace{-32pt}
    \caption{
Overall framework of MobTA. 
MobTA constructs three task vectors corresponding to mobility trajectory generation in the source city, as well as bus-timetable-based trajectory generation in both the source and target cities. 
It then models the parameter shift from bus-timetable-based  trajectory generation to mobility trajectory generation in the source city and applies this shift to the target city’s bus-timetable-based trajectory generation task vector, enabling mobility trajectory generation for the target city.
    }
    \label{fig:model}
    \vspace{-5pt}
\end{figure*}

As shown in Table~\ref{sh bus 2015}, bus stop sequences exhibit a high degree of consistency over time. 
When the analysis is restricted to area within Outer Ring Road (the  study area of this work), the similarity becomes even more pronounced.
These results indicate that bus station sequences 
preserve a stable spatial structure over long time span.
Detailed experimental settings are provided in Appendix~\ref{sec:detail spatial sta}.

\subsubsection{Spatiotemporal Correlations Between Bus Timetables and Mobility Trajectories}

After establishing the spatial stability of bus station sequences, a more critical question is whether bus timetables exhibit meaningful spatiotemporal correlations with mobility trajectories.
If bus timetables only capture spatial structures weakly related to mobility patterns, their stability alone would be insufficient to support subsequent modeling. 
Therefore, we further examine the spatiotemporal relationships between bus timetables and mobility trajectories.

\begin{table}[ht]
  \centering
  \caption{Spatiotemporal Relationship Between Bus Timetables and Mobility Trajectories}
  \vspace{-2mm}
  \label{tab:spatial sim}
  \vskip 0.1in
     \resizebox{\columnwidth}{!}{%
    \setlength{\tabcolsep}{3.5pt} 
    \begin{tabular}{l|ccccc|c}
      \toprule
      City &$\mathrm{JSD}_c$($\downarrow$) & Cos($\uparrow$) & Pear-S($\uparrow$) & PCov($\uparrow$) & RCov($\uparrow$) & Pear-T($\uparrow$)\\
      \midrule
      SH  & 0.1142 & 0.7595 & 0.6592 & 86.86 & 86.87 & 0.8180\\
      WX      & 0.2064 & 0.6595 & 0.6077 & 55.71 & 51.42 & 0.9180\\
      
      SG & 0.0802 & 0.8726 & 0.8457 & 80.08 & 63.82 & 0.9702 \\
      \bottomrule
    \end{tabular}}
  \vskip -0.1in
\end{table}

From a spatial perspective, Table~\ref{tab:spatial sim} reports  comparisons between bus timetables and mobility trajectories.
The results show that bus timetables and mobility trajectories exhibit relatively high consistency under Jensen–Shannon divergence, cosine similarity, and spatial Pearson correlation, indicating  overlap in major regions. 
However, these metrics are lower than those under complete alignment, suggesting that bus timetables and  mobility trajectories do not share identical spatial patterns. 
Regarding point and route coverage, while a large proportion of mobility trajectories spatially overlap with bus timetable coverage, a non-negligible fraction of mobility activities occur outside this coverage.

From a temporal perspective,  Table~\ref{tab:spatial sim} further reports the temporal Pearson correlation between bus timetables and mobility trajectories for each city, with their specific hourly distributions depicted in Fig.\ref{fig:sh tem dis}-\ref{fig:wx tem dis} in Appendix.~\ref{sec:bus timetable relationship detail}.
The results indicate that bus timetables and mobility trajectories exhibit similar flow rhythms, particularly during daytime hours. 
In late-night periods when bus services are sparse, mobility trajectories remain relatively active, leading to noticeable deviations.

Overall, although bus timetables cannot fully capture urban mobility behaviors, they remain  correlated with mobility trajectories in spatial and temporal dimensions.
These empirical findings suggest that bus timetables can serve as an effective structural reference for understanding urban mobility, providing practical support for trajectory generation where target city mobility trajectories are inaccessible. 
Detailed experimental settings are provided in Appendix.~\ref{sec:bus timetable relationship detail}.

\section{Method}
\subsection{Framework Overview}

Motivated by the empirical observations in Sec.\ref{sec:emp}, we establish the intuition of MobTA: 
bus timetables reveal a stable mobility 'backbone' in terms of spatial layout and temporal regularity, while mobility trajectories provide more flexible and fine-grained mobility patterns that complement this backbone.
Their discrepancy exhibits a certain degree of transferability across
cities.
When target city mobility trajectories are inaccessible, directly learning a mobility trajectory generation model is infeasible. Moreover, training solely on bus timetables tends to overfit bus-specific patterns and fails to generalize.
A more viable perspective is to explicitly model the systematic discrepancy between bus-timetable-based trajectory generation and mobility trajectory generation, and to treat this discrepancy  as transferable knowledge.
Based on this idea, MobTA, as shown in Fig.~\ref{fig:model},  leverages task arithmetic in the parameter space to transfer
such knowledge across cities, enabling mobility trajectory generation under the bus-conditioned zero-shot setting.

\subsection{Task Vectors for Trajectory Generation}
To apply effective task arithmetic to trajectory generation, it is critical to ensure that task vectors are aligned in parameter space, such that vector arithmetic corresponds to consistent functional edits rather than representation mismatches.
Therefore, we derive task vectors for each trajectory generation task by fine-tuning under a unified trajectory representation and serialization scheme.

\subsubsection{Unified Trajectory Representation and Serialization}
\label{sec:traj repre ser}
We adopt a unified spatial discretization and tokenization strategy by partitioning the study region of each city into regular grids with the same spatial resolution.
We designate the grid containing  geometric center of the study region as the anchor grid, indexed as
\texttt{<LOC\_0>}.
Grid indices are then assigned to surrounding grids using a spiral indexing strategy that expands outward from the center in a clockwise manner (see Fig.~\ref{fig:grid} in Appendix~\ref{sec:spatial app}).
Crucially, this indexing strategy enables different cities to share the same set of location tokens based on the relative positions of grids within each city.
For instance, \texttt{<LOC\_0>}
consistently represents the center, while larger indices correspond to more peripheral regions.
By incorporating these tokens into a shared vocabulary, we ensure that location-related embeddings are comparable across cities, which is a prerequisite for valid cross-city task vector arithmetic. This design ensures structural comparability without assuming strict geometric alignment, enabling task vector arithmetic to generalize across cities by focusing on the systematic discrepancy between bus timetables and mobility patterns, rather than city-specific spatial layouts.

To enable trajectory generation, we serialize each trajectory into a textual sequence.
Each record is formatted as 
“arrival time is $t_i$, location is \texttt{<LOC\_i>} \#”, and the entire trajectory is concatenated sequentially.
Here $t_i$ is a
 numerical token representing the time-of-day.
We formulate trajectory generation as a standard autoregressive language modeling task, trained via next-token prediction.
This allows the model to learn spatiotemporal  patterns directly into its parameters.

\subsubsection{Task Vector Construction}
Under the unified trajectory representation and serialization, we obtain task vectors by fine-tuning LLMs on trajectory generation tasks  over bus timetables and mobility trajectories.
Beyond attention modules, we jointly consider parameter shifts in the embedding layer, the output head, and feed-forward networks as components of task vector. Together, these parameter shifts capture complementary aspects of trajectory generation behavior and form a unified representation of task-specific adaptations.

For embedding  and output layers, we initialize the parameters of newly added spatial tokens using the global mean of pre-trained vocabulary embeddings (see Appendix~\ref{sec:repro} and \ref{sec:automa} for details), and perform full-parameter fine-tuning. 
Unlike Geollama~\cite{li2025geo}, which encodes locations as numerical identifiers that are tokenized into multiple sub-tokens, our approach assigns a dedicated token to each spatial grid.
This avoids fragmented parameter updates across sub-tokens lacking explicit spatial semantics, and yields coherent and composable task vectors.

For LLM backbone, we employ LoRA to adapt both the Q, K, and V projections in attention layers and the up, down, and gate projections in feed-forward networks.
This design stems from the insight that trajectory generation relies not only on modelling sequential dependencies, but also  on the nonlinear transformations  to shape spatiotemporal patterns~\cite{dong2021attention,geva2022transformer}. 

We define the task vector for a specific trajectory generation task as the parameter difference between the fine-tuned model and the pretrained model. 
To reflect the distinct functional roles, we decompose task vector $\mathbf{V}$ into three sub task vectors: $\{\mathbf{V}_E,\mathbf{V}_L,\mathbf{V}_H\}$, corresponding to the embedding layer, LoRA modules, and output head, respectively.

\subsection{Task Arithmetic for Bus-Guided Zero-shot Trajectory Generation}
Given the task vectors constructed for different trajectory generation tasks, we  construct the mobility trajectory generation task vector for target city via task arithmetic.
We assume that the parameter shift from bus-timetable-based trajectory generation to  mobility trajectory generation is relatively stable, and can  be transferred across cities.
Let $\mathbf{V}^{src-bus}$ and $\mathbf{V}^{src-mob}$ denote the task vectors for bus timetable and  mobility trajectory generation in source city, and let $\mathbf{V}^{tgt-bus}$ 
 denote the task vector for bus-timetable-based trajectory generation in target city. 
 By applying parameter shift in source city to $\mathbf{V}^{tgt-bus}$, we construct the mobility trajectory generation task vector for  target city:
\begin{equation}
\hat{\mathbf{V}}_E^{\text{tgt-mob}} = \mathbf{V}_E^{\text{tgt-bus}} + \mu_E \left(  \mathbf{V}_E^{\text{src-mob}}-\mathbf{V}_E^{\text{src-bus}}  \right),
\end{equation}
\begin{equation}
\hat{\mathbf{V}}_L^{\text{tgt-mob}} = \mathbf{V}_L^{\text{tgt-bus}} + \mu_L \left( \mathbf{V}_L^{\text{src-mob}}-\mathbf{V}_L^{\text{src-bus}}  \right),
\end{equation}
\begin{equation}
\hat{\mathbf{V}}_H^{\text{tgt-mob}} = \mathbf{V}_H^{\text{tgt-bus}} + \mu_H \left( \mathbf{V}_H^{\text{src-mob}}-\mathbf{V}_H^{\text{src-bus} } \right),
\end{equation}
where $\mu_E,\mu_L$, and $\mu_H$ 
control the magnitude of  transferred parameter shift within each subspace. 
The resulting sub task vectors are added to the pre-trained model parameters to obtain a model for target-city mobility trajectory generation.
This process does not involve any further gradient based optimization and does not require access to real mobility trajectories from target city, thereby providing an effective solution for bus-guided zero-shot trajectory generation.

\subsection{Theoretical Analysis of MobTA's Stability under Base and Instruction-Tuned LLMs}
\label{sec:proof}
Prior studies on task arithmetic mainly consider task vectors derived from base LLMs~\cite{kawamoto2025cross,gonzalez2025fairness}, while those  from instruction-tuned models are less explored. 
A common concern is that instruction tuning introduces additional parameter differences related to instruction following and safety alignment, which may interfere with task-specific parameter changes and  undermine linear additivity of task vectors~\cite{chen2025layer}.

However, we argue that these concerns are mitigated in MobTA.
Trajectory generation in MobTA does not rely on complex instruction understanding or multi-turn interactions; instead, differences among trajectory generation tasks are primarily reflected in newly introduced spatial tokens within fixed structured trajectory sequences.
Under this setting, task arithmetic applied to task vectors obtained from either base or instruction-tuned LLMs exhibits comparable performance on trajectory generation.
This consistency arises because, in our setting, trajectory generation activates a highly constrained task-relevant parameter subspace, while the  parameter differences introduced by instruction tuning have only a small projection onto this subspace.

\subsubsection{Task-Relevant Parameter Subspace}
We first formalize which parameter directions  are truly influential for trajectory generation in our setting.
Let $\theta$ denote the model parameters. 
Given a serialized  trajectory sequence 
$x$, as detailed in Sec.~\ref{sec:traj repre ser}, the logit for predicting token $T$ is denoted as $z_T(\theta;x)$.
In MobTA,  trajectory generation is formulated as predicting trajectory-related tokens $T \in \mathcal{V}_{\mathrm{traj}}$, consisting of newly introduced
spatial location tokens and time-related numerical tokens.
We define the task-relevant parameter subspace for MobTA's trajectory generation, denoted as $\mathcal{T}_{\mathrm{traj}}$, as parameter directions spanned by the gradients associated with trajectory-related tokens:
\begin{equation}
    \small
    \mathcal{T}_{\mathrm{traj}}
\;:=\;
\operatorname{span}
\left\{
\nabla_{\theta} z_{T}(\theta; x)
\;\middle|\;
x \sim \mathcal{X}_{\mathrm{traj}},
\;
T \in \mathcal{V}_{\mathrm{traj}},\theta=\theta_0^{(B)}
\right\},
\end{equation}
where $\mathcal{X}_{\mathrm{traj}}$ is the distribution of serialized  trajectory sequences, and $\theta_0^{(B)}$  denotes the base LLM initialization,
which serves as a fixed reference point for defining the task-relevant subspace.
We further denote $\mathbf{P}_{\mathrm{traj}}$ as the orthogonal projection operator onto the subspace $\mathcal{T}_{\mathrm{traj}}$.

Although each gradient vector $\nabla_{\theta} z_{T}(\theta; x)$ has the same dimensionality as the full parameter vector $\theta$, the trajectory generation  in our setting induces 
intrinsic sparsity and structural constraints.
Parameter directions associated with instruction following or safety alignment typically yield negligible gradients.
In contrast, significant gradient magnitudes are  localized to parameter directions associated with trajectory-specific tokens.
Thus, $\mathcal{T}_{\mathrm{traj}}$ is expected to be low-dimensional and highly focused on trajectory generation.

\subsubsection{Projection of base and instruction-tuned LLMs onto Task-Relevant Subspace}

Let $\theta_0^{(B)}$ and $\theta_0^{(I)}$ denote the initialization parameters of the base and the instruction-tuned LLM, and define their difference as $\delta_0:=\theta_0^{(I)}-\theta_0^{(B)}$.
Although $\delta_0$ may be substantial in the full parameter space, instruction tuning primarily affects capabilities related to instruction following and safety alignment,
which are typically associated with generic language-level behaviors rather than trajectory generation.
In MobTA, trajectories are serialized using a fixed structured input format and trajectory information is primarily encoded through newly introduced spatial tokens.
This motivates the following  assumption: the initialization difference $\delta_0$ induced by instruction tuning has a relatively small projection onto subspace $\mathcal{T}_{\mathrm{traj}}$, namely
\begin{equation}
\frac{\| P_{\mathrm{traj}} \, \delta_0 \|}{\| \delta_0 \|} \le \epsilon, \quad \epsilon \ll 1 .
\end{equation}
Under this assumption, despite potentially large global parameter shifts, the base and instruction-tuned models remain close within the task-relevant subspace, which helps explain the stability of task arithmetic operations in MobTA.

\subsubsection{Projection Stability of Task Vectors}
\label{first order}
For a trajectory dataset $\mathcal{D}$ and a given model's initialization $\theta_0^{(\cdot)},(\cdot) \in \{B,I\}$, we define the corresponding task vector as the parameter shift obtained via fine-tuning:
\begin{equation}
    \mathbf{V}(\theta_0^{(\cdot)};\mathcal{D})=\theta^*(\theta_0^{(\cdot)};\mathcal{D})-\theta_0^{(\cdot)},
\end{equation}
where $\theta^*(\theta_0^{(\cdot)};\mathcal{D})$ is the  parameters after finetuning.
Under a small-step, parameter-efficient fine-tuning strategy, this parameter shift admits a standard first-order approximation:
\begin{equation}
   \mathbf{V}(\theta_0; \mathcal{D}) \approx -\eta \nabla_{\theta} \mathcal{L}(\theta_0; \mathcal{D}),
\end{equation}
where $\mathcal{L}(\theta_0; \mathcal{D})$ is the training loss for trajectory generation and $\eta$ denotes the effective step size.
For dataset $D$, we consider the projection of the task vector difference induced by base and instruction-tuned models onto subspace $\mathcal{T}_{traj}$:

\begin{equation}
\begin{aligned}
&\mathbf{P}_{\mathrm{traj}}\bigl(\mathbf{V}_B(\mathcal{D}) - \mathbf{V}_I(\mathcal{D})\bigr) \approx \\
&\quad -\eta \,\mathbf{P}_{\mathrm{traj}}\bigl(\nabla_{\theta} \mathcal{L}(\theta_0^{(B)}; \mathcal{D}) - \nabla_{\theta} \mathcal{L}(\theta_0^{(I)}; \mathcal{D})\bigr)
\end{aligned}
\end{equation}
Under a first-order approximation, this projected  difference is controlled by the small projection of $\delta_0$ onto $\mathcal{T}_{\mathrm{traj}}$ and the weak interaction between $\mathcal{T}_{\mathrm{traj}}$ and its orthogonal complement, and thus remains numerically small.
See Appendix~\ref{sec:detail theory} for more details.

\subsubsection{Stability of MobTA Task Arithmetic}
MobTA constructs the target mobility trajectory generation task vector $\hat{\mathbf{V}}^{\text{tgt-mob}}$ via task arithmetic (Eq.2-4).
Within subspace $\mathcal{T}_{\mathrm{traj}}$, the difference  between $\hat{\mathbf{V}}^{\text{tgt-mob}}$ obtained from base and instruction-tuned LLMs  satisfies
{\fontsize{8.5pt}{8pt}\selectfont
\begin{equation}
\begin{aligned}
\small
\mathbf{P}_{\mathrm{traj}}
\bigl(
\hat{\mathbf{V}}_B^{\text{tgt-mob}} - \hat{\mathbf{V}}_I^{\text{tgt-mob}}
\bigr)
&=
\mathbf{P}_{\mathrm{traj}}
\bigl(
\mathbf{V}_B(\mathcal{D}^{bus}_{c_{tgt}}) - \mathbf{V}_I(\mathcal{D}^{bus}_{c_{tgt}})
\bigr)
\\
&\quad +
\mu \Bigl(
\mathbf{P}_{\mathrm{traj}}
\bigl(
\mathbf{V}_B(\mathcal{D}^{mob}_{c_{src}}) - \mathbf{V}_I(\mathcal{D}^{mob}_{c_{src}})
\bigr)
\\
&\quad\quad-
\mathbf{P}_{\mathrm{traj}}
\bigl(
\mathbf{V}_B(\mathcal{D}^{bus}_{c_{src}}) - \mathbf{V}_I(\mathcal{D}^{bus}_{c_{src}})
\bigr)
\Bigr).
\end{aligned}
\end{equation}
}
As each term on the right-hand side is small within $\mathcal{T}_{\mathrm{traj}}$, their linear combination remains small within this subspace.  
The logits of trajectory-related tokens are primarily influenced by parameter shifts within $\mathcal{T}_{\mathrm{traj}}$, and insensitive to parameter shifts in its orthogonal complement $\mathcal{T}_{\mathrm{traj}}^{\perp}$. 
When $\|\mathbf{P}_{\mathrm{traj}}
\bigl(
\hat{\mathbf{V}}_B^{\text{tgt-mob}} - \hat{\mathbf{V}}_I^{\text{tgt-mob}}
\bigr)\|$ is small, the distributions over location and time tokens at each generation step remain close.
Through the autoregressive generation process, this step-wise proximity propagates to the distribution over complete trajectories, which explains the stability of MobTA task arithmetic across base and instruction-tuned LLMs.

\section{Experiment}
 We conduct extensive experiments under the problem setting of bus-conditioned zero-shot trajectory generation, where real  mobility trajectories in the target city are completely inaccessible during training.
 We focus on trajectory generation performance, behavior under varying spatial data coverage, and robustness across LLM architectures.
 Ablation studies and hyperparameter sensitivity analysis are detailed in Appendix~\ref{ablation} and \ref{hyper}.
Code is available at \url{https://anonymous.4open.science/r/MobTA-2370/}.
Furthermore, to enhance transparency and reproducibility, we report practical implementation considerations and technical nuances in Appendix~\ref{consider}.

\begin{table}[ht]
  \centering
  \caption{Distributional discrepancy (JSD) between bus timetables and real mobility trajectories.}
  \label{table:jsd_updated}
  \vskip 0.1in
  \begin{small}
    \setlength{\tabcolsep}{6pt} 
    \begin{tabular}{lcccc}
      \toprule
      City & Dist. & Rad. & Dur. & Loc. \\
      \midrule
      Shanghai (SH)  & 0.1765 & 0.1777 & 0.0374 & 0.1643 \\
      Wuxi (WX)      & 0.3055 & 0.2841 & 0.0379 & 0.2483 \\
      Singapore (SG)        & 0.1747 & 0.1356 & 0.0361 & 0.0930 \\
      \bottomrule
    \end{tabular}
  \end{small}
  \vskip -0.1in
\end{table}

\begin{table*}[t]
\centering
\small
\caption{ Trajectory generation performance measured by JSD between generated and real mobility trajectories.
Smaller values indicate better performance.
Geollama$^+$ (mob) here stands for the upper bound which is finetuned using real target city mobility trajectories.}
\setlength{\tabcolsep}{3pt}
\begin{tabular}{lcccccccccccc}
\toprule
& \multicolumn{4}{c}{\textbf{SH}} 
& \multicolumn{4}{c}{\textbf{WX}} 
& \multicolumn{4}{c}{\textbf{SG}} \\
\cmidrule(lr){2-5} \cmidrule(lr){6-9} \cmidrule(lr){10-13}
\textbf{Method} 
& Dist. & Rad. & Dur. & Loc. 
& Dist. & Rad. & Dur. & Loc.
& Dist. & Rad. & Dur. & Loc. \\
\midrule

TimeGeo                     
& 0.3555 & 0.4009 & 0.3023 & 0.3150 
& 0.2955 & 0.2881 & 0.2260 & 0.2062 
& 0.4325 & 0.4817 & 0.1810 & 0.3856 \\

MoveSim                     
& 0.2789 & 0.3954 & 0.1161 & 0.1408 
& 0.3857 & 0.3552 & 0.1596 & 0.1118 
& 0.2892 & 0.3171 & 0.0700 & 0.2317 \\

TSG                         
& 0.3149 & 0.2763 & 0.0912 & 0.1800 
& 0.3969 & 0.2656 & 0.1506 & 0.2059 
& 0.3306 & 0.3928 & 0.0633 & 0.3357 \\

TrajGDM                     
& 0.2238 & 0.1913 & 0.0512 & 0.1067 
& 0.2591 & 0.1844 & 0.0837 & 0.0901 
& 0.2355 & 0.2769 & 0.0351 & 0.1864 \\

DiffTraj                    
& 0.2763 & 0.2124 & 0.0673 & 0.1156 
& 0.3209 & 0.2149 & 0.0748 & 0.1742 
& 0.2951 & 0.3080 & 0.0477 & 0.2156 \\

TrajGen                     
& 0.2880 & 0.2381 & 0.0692 & 0.1824 
& 0.3701 & 0.3029 & 0.1155 & 0.2093 
& 0.3074 & 0.3347 & 0.0489 & 0.3389 \\

Traveller                   
& 0.2167 & 0.1746 & 0.0615 & 0.0980 
& 0.2726 & 0.1730 & 0.1021 & 0.1366 
& 0.2325 & 0.2476 & 0.0433 & 0.2120 \\

Geollama$^{+}$ (bus)                
& 0.2058 & 0.1493 & 0.0342 & 0.1175 
& 0.2936
 & 0.1684
 & 0.0917
 & 0.0997
& 0.2193
 & 0.2692
 & 0.0471
 & 0.2416
 \\

\midrule

COLA (SH$\rightarrow$)                   
& \multicolumn{1}{c}{--} & \multicolumn{1}{c}{--} & \multicolumn{1}{c}{--} & \multicolumn{1}{c}{--}
& 0.1651 & 0.1308 & 0.0529 & 0.1056 
& 0.2170 & 0.1844 & 0.0436 & 0.1193 \\

COLA (WX$\rightarrow$)                   
& 0.1831 & 0.1564 & 0.0344 & 0.1544 
& \multicolumn{1}{c}{--} & \multicolumn{1}{c}{--} & \multicolumn{1}{c}{--} & \multicolumn{1}{c}{--}
& 0.2539 & 0.2161 & 0.0713 & 0.1842 \\

COLA (SG$\rightarrow$)                   
& 0.1610 & 0.1497 & 0.0423 & 0.1364 
& 0.1920 & 0.1544 & 0.0587 & 0.1101
& \multicolumn{1}{c}{--} & \multicolumn{1}{c}{--} & \multicolumn{1}{c}{--} & \multicolumn{1}{c}{--} \\

MobTA (SH$\rightarrow$)     
& \multicolumn{1}{c}{--} & \multicolumn{1}{c}{--} & \multicolumn{1}{c}{--} & \multicolumn{1}{c}{--}
& 0.0952 & 0.0576 & \textbf{0.0472} & 0.0664 
& \textbf{0.0855} & \textbf{0.0469} & \textbf{0.0387} & \textbf{0.0961} \\

MobTA (WX$\rightarrow$)     
& 0.0969 & 0.0639 &  \textbf{0.0206} & \textbf{0.0487}

& \multicolumn{1}{c}{--} & \multicolumn{1}{c}{--} & \multicolumn{1}{c}{--} & \multicolumn{1}{c}{--}
& 0.1144 & 0.0656 & 0.0417 & 0.1289 \\

MobTA (SG$\rightarrow$)     
& \textbf{0.0749} & \textbf{0.0374} & 0.0311 & 0.0552 
& \textbf{0.0847}
 & \textbf{0.0539}
 & 0.0508
 & \textbf{0.0597}
& \multicolumn{1}{c}{--} & \multicolumn{1}{c}{--} & \multicolumn{1}{c}{--} & \multicolumn{1}{c}{--} \\

\midrule

Geollama$^{+}$ (mob)        
& 0.0515 & 0.0284 & 0.0192 & 0.0293 
& 0.0487
 & 0.0372
 & 0.0286
 & 0.0387
& 0.0715
 & 0.0877
 & 0.0312
 & 0.0577
 \\

\bottomrule
\end{tabular}
\label{tab:cross_city_jsd}
\end{table*}

\subsection{Experimental Setup}
We conduct experiments on three real-world urban datasets: Shanghai (SH), Wuxi (WX), and Singapore (SG). We compare MobTA with  cross-city transfer method COLA and several representative trajectory generation models.
Experiments are conducted on four NVIDIA RTX A5000 GPUs.
Unless specified, the LLM used is LLaMA-3.2-3B.
Detailed dataset descriptions and processing procedure are provided in Appendix~\ref{app data}.
Baseline details can be seen in Appendix~\ref{baseline detail}, while  evaluation metrics  are detailed in Appendix~\ref{metric detail}.

\subsection{Trajectory Generation Performance Comparison}
Table~\ref{table:jsd_updated}  shows the distributional differences between bus timetables and real mobility trajectories in each city.
Discrepancies exist in distance, radius, and TrajLoc, further indicating that bus timetables alone are insufficient to capture urban mobility behavior.

Table~\ref{tab:cross_city_jsd} compares the trajectory generation ability of MobTA and baselines.
Non-transfer methods, which are trained solely on bus timetables, are
inherently constrained by the coverage and statistics of bus timetables, and
thus fail to bridge the distributional gap observed in Table~\ref{table:jsd_updated}.
COLA is restricted to generating trajectories within locations observed
during training, i.e., those covered by bus timetables. 
This limits its generalization to uncovered areas, leading to deficiencies in location-related metrics.
These results show that existing baselines are not designed to handle bus-conditioned zero-shot trajectory generation.
They are ineffective for  cities where real mobility trajectories are inaccessible.
In contrast, MobTA achieves consistently better performance.
Its overall generation quality  approaches that of models  fine-tuned on target city mobility trajectories. 
These results 
indicate that MobTA can  effectively capture the overall mobility patterns of the target city, even in the absence of real mobility data.

\begin{table*}[t]
\centering
\small
\caption{MobTA’s trajectory generation behavior under  different spatial data coverage conditions.}
\setlength{\tabcolsep}{4pt}
\renewcommand{\arraystretch}{1.05}
\begin{tabular}{cccccccccc}
\toprule
\textbf{WX Mob} & \textbf{WX Bus} & \textbf{SH Bus} 
& \textbf{\#Loc} & \textbf{JSD$_{dur}$} 
& $\rho_{\text{traj}}^{\text{real}}$ 
& $\rho_{\text{traj}}^{\text{gen}}$ 
& $\rho_{\text{visit}}^{\text{real}}$ 
& $\rho_{\text{visit}}^{\text{gen}}$ \\
\midrule

$\times$ & $\times$ & $\times$ 
& 70  & 0.0176 
& 0.0521 & 0.0347 & 0.0084 & 0.0019 \\

$\times$ & $\times$ & $\checkmark$ 
& 140 & 0.0378 
& 0.2649 & 0.3361 & 0.1112 & 0.1323 \\

$\times$ & $\checkmark$ & $\times$ 
& 0   & --     
& -- & -- & -- & -- \\

$\times$ & $\checkmark$ & $\checkmark$ 
& 2   & 0.2399 
& 0.0568 & 0.0197 & 0.0081 & 0.0015 \\

$\checkmark$ & $\times$ & $\times$ 
& 15  & 0.0368 
& 0.0053 & 0.0245 & 0.0007 & 0.0013 \\

$\checkmark$ & $\times$ & $\checkmark$ 
& 135 & 0.0254 
& 0.4767 & 0.6332 & 0.1457 & 0.2456 \\

$\checkmark$ & $\checkmark$ & $\times$ 
& 18  & 0.0104 
& 0.0309 & 0.0228 & 0.0038 & 0.0012 \\

$\checkmark$ & $\checkmark$ & $\checkmark$ 
& 350 & 0.0188 
& 0.8820 & 0.8756 & 0.7220 & 0.6148 \\

\bottomrule
\end{tabular}

\label{tab:source_combination_ratio}
\end{table*}

\subsection{Trajectory Generation Behavior under Different Spatial Data Coverage Conditions}
We analyze MobTA’s trajectory generation behavior under  different spatial data coverage conditions.
Taking the WX $\rightarrow$ SH transfer setting as a case study example,  we categorize Shanghai locations based on whether they are covered by WuXi mobility trajectories, Wuxi  bus timetables, and Shanghai bus timetables.
Fig.~\ref{fig:shanghai grids} in Appendix~\ref{detail: data coverage} visualizes these spatial data coverage categories.
We evaluate  the generation behavior in terms of stay duration distribution, spatial reachability, and visit frequency.
 Details of the evaluation are provided in Appendix~\ref{detail: data coverage}.
Table~\ref{tab:source_combination_ratio} 
shows MobTA’s performance under different spatial data coverage conditions.
Overall, MobTA exhibits stable trajectory generation behavior, with particularly strong performance when locations are covered by all three data sources.

\begin{table}[t]
\centering
\caption{Performance  of different LLMs on WX $\rightarrow$ SH.}
\label{tab:diff LLM}
\resizebox{\columnwidth}{!}{
\begin{tabular}{lcccc}
\hline
Model &Dist. & Rad. & Dur. & Loc. \\
\hline
LLaMA-3.2-3B        & 0.0969 & 0.0639 & 0.0206 & 0.0487 \\
Qwen3-1.7B-Base     & 0.0846 & 0.0645 & 0.0208 & 0.0662 \\
\midrule
LLaMA-3.2-3B-Instruct        & 0.0926 & 0.0601 & 0.0231 & 0.0496 \\
Qwen3-1.7B          & 0.0830 & 0.0628 & 0.0198 & 0.0532 \\
\hline
\end{tabular}
}
\vspace{-20pt}
\end{table}

\subsection{Robustness across LLM Architectures}
Table~\ref{tab:diff LLM} compares the performance of MobTA  across different LLM families and their respective base and instruction-tuned variants under the WX $\rightarrow$ SH transfer setting. 
MobTA maintains consistent generation quality on both LLaMA and Qwen, indicating that the proposed task arithmetic approach is not tied to a specific LLM architecture.
Furthermore, within each model family, base and instruction-tuned variants exhibit only negligible performance differences.
This empirical observation aligns with the theoretical analysis in Sec.~\ref{sec:proof}, which shows that, for trajectory generation, the difference between task vectors derived from base and instruction-tuned LLMs is marginal when projected onto the task-relevant parameter subspace.
As a result, MobTA's trajectory generation ability is consistent across base and instruction-tuned LLMs.

\section{Conclusion}
This paper introduces bus-conditioned zero-shot trajectory generation, a challenging yet practical new problem setting in which no real mobility trajectories from the target city are accessible, and trajectory generation relies solely on  bus timetables in the target city and source city data.
Under this setting, we propose MobTA, the first approach to  apply task arithmetic to trajectory generation.
By  modeling the parameter shift from bus-timetable-based trajectory generation to mobility trajectory generation in the source city, and applying this shift to target city through arithmetic operations on task vectors, MobTA enables  trajectory generation that closely matches real mobility patterns in the target city.
From a theoretical perspective, we  analyze the stability of MobTA across base and instruction-tuned large language models.
Extensive experiments demonstrate MobTA's great superiority.
As future work, it would be valuable to  investigate the relationship between task arithmetic scaling coefficients $\mu_E,\mu_L,\mu_H$ and data characteristics like distributional discrepancies between source and target cities or gradient norms of task vectors.
Moreover, extending MobTA to cities with  underdeveloped public bus systems is an important direction for future research.
In addition, exploring whether explicit bus stop attributes like station names or topological relationships can further enhance trajectory generation performance remains an interesting direction.

\newpage

\bibliography{example_paper}
\bibliographystyle{icml2026}

\newpage
\appendix
\section{Detailed related work}

\subsection{Trajectory Generation}
\label{sec:detailed traj}
Human mobility trajectory data  encapsulate rich spatiotemporal semantic information and serve as a fundamental data source for understanding urban dynamics and human interactions~\cite{chen2024deep}.
Such data play a critical role across a wide range of smart city applications~\cite{liu2024nextlocllm,ju2025trajllm}.
In urban planning, analyses of large-scale mobility trajectories enable the characterization of commuting patterns and  travel demands, which supports decisions on infrastructure service arrangement, land-use planning, and establishment of many other urban-related policies~\cite{zhou2024large,ke2026uncovering}.
In mobility prediction, historical trajectory data constitute a key input for traffic flow modeling and are widely used for congestion forecasting, estimated time of arrival prediction, and personalized route recommendation~\cite{tang2024instruction,anda2024personalized,liu2025mixture,chen2025enhancing}.
Moreover, for epidemic control, individual mobility trajectories provide essential information for tracing transmission chains, quantifying contact risks, and evaluating the effectiveness of intervention measures, thereby playing a crucial role in mitigating disease spread~\cite{kim2021impact,hu2021human,elarde2021change}.

Despite their substantial practical value, the acquisition and sharing of real-world mobility trajectory data are highly challenging~\cite{chen2025trajectory}.
On the one hand, trajectory data are typically collected by companies like ride-hailing platforms and telecommunication operators. 
Due to commercial restrictions, they are rarely open to the public.
 On the other hand, trajectory data inherently contain highly identifiable spatiotemporal behavior patterns. 
 Releasing such data can result in severe privacy risks. 
 These challenges jointly limit the accessibility of high-quality mobility trajectories.

Consequently, trajectory generation has emerged as an effective solution~\cite{kulkarni2017generating,ouyang2018non}. 
Its core objective is to capture the underlying spatiotemporal  patterns and human mobility behaviors from real mobility trajectory data, and to synthesize trajectories accordingly. 
These synthetic trajectories aim to preserve the statistical properties and behavioral characteristics of real mobility data while not corresponding to any real individual, thereby reducing commercial constraints and privacy risks, as well as  enabling downstream urban tasks. 
Existing trajectory generation approaches can majorly be categorized into model-based methods and learning-based methods~\cite{chen2025trajectory,zhu2025trajectory}.

\subsubsection{Model-based Trajectory Generation}
Model-based trajectory generation approaches are built on the assumption that human mobility behaviors can be characterized by predefined statistical patterns or behavioral rules.
These methods typically employ explicit probabilistic models or handcrafted mechanisms to describe transitions between mobility states. 
Because of their interpretability and low data requirements, model-based approaches were widely studied in the early era.
SLAW~\cite{lee2009slaw}  assumes that people prefer to visit spatially proximate locations. 
It models both travel distances and dwell times using power-law distributions, thereby reproducing the heavy-tailed statistics commonly observed in real trajectories.
SWIM~\cite{mei2009swim} adopts a social behavior perspective and models mobility decisions as a weighted selection over locations, where the weights are influenced by both location popularity and distance to residents' home. 
SIMPS~\cite{borrel2008simps}  incorporates sociological theories by distinguishing between social and isolated behaviors.
It models interpersonal relationships through social graphs, leading to power-law distributions in contact duration.
\cite{song2010modelling}  combines exploration of new locations with preferential return to previously visited ones, so as to capture the dynamic balance between new place seeking and habitual behavior.
Building upon this framework, \cite{pappalardo2015returners} incorporates gravity models that jointly consider spatial distance and location importance, enabling fine-grained modeling of decision preferences.

In addition, several methods use  hidden markov models.
\cite{mathew2012predicting} and \cite{bindschaedler2016synthesizing} employ hidden Markov models to capture latent state transitions in mobility trajectories, enabling next location prediction and trajectory generation. 
 \cite{baratchi2014hierarchical} introduces hierarchical hidden semi-Markov models, which enhance the expressiveness of temporal dynamics while preserving model interpretability.

\subsubsection{Learning-based Trajectory Generation}
Learning-based trajectory generation methods aim to model and synthesize complex human mobility patterns by directly learning underlying spatiotemporal patterns from real trajectory data. 
In contrast to model-based approaches, these methods do not rely on strong predefined assumptions, but instead leverage deep learning models to automatically capture high-dimensional correlations and nonlinear dependencies inherent in mobility trajectory data.
As a result, learning-based methods generally achieve superior generation quality. 
Existing learning-based methods are primarily based on  generative adversarial networks, variational autoencoders, diffusion models, and more recently, large language models.

Generative adversarial networks (GANs) are among the earliest deep generative models adopted for trajectory generation. 
In a typical GAN framework, a generator is trained to generate trajectories that resemble real mobility data, while a discriminator aims to distinguish between real and generated trajectories.
The two components are jointly optimized through adversarial training. 
\cite{feng2020learning} incorporates human mobility priors into the GAN framework by combining self-attention-based temporal modeling modules with region-level spatial networks. 
\cite{cao2021generating}  explicitly balances the realism of generated trajectories with their utility for downstream tasks during adversarial training.
LSTM-TrajGAN~\cite{rao2020lstm} integrates LSTMs into the GAN framework to preserve both temporal consistency and privacy.
In scenarios with strict road network constraints, \cite{wang2021large} proposes a two-stage generation strategy, where global trajectory distributions are first learned at a coarse grid level and then refined into fine-grained paths under road network constraints.
 STAGE~\cite{cao2025stage} incorporates Transformers  into the GAN framework and adopts multi-task learning to better capture long-range spatiotemporal dependencies.

 Variational autoencoders (VAEs)  typically assume that observed trajectories are generated from underlying latent variables and learn mappings between trajectory observations and latent representations through encoder–decoder architectures. 
\cite{ding2019multi} proposes a generative model for multi-vehicle trajectory generation.
It employs a bidirectional GRU encoder and a multi-branch decoder to jointly model the motion states of interacting vehicles in complex traffic scenarios. 
\cite{huang2019variational} combines VAEs with a Seq2Seq framework and introduces latent variable modeling at each time step, enabling continuous trajectory generation. 
STULIG~\cite{zhou2020toward} incorporates a Gaussian mixture prior to capture the multimodal nature of trajectory data, and adopts convolutional encoders and decoders to improve the representation of complex spatiotemporal patterns.
Compared to GAN-based models, VAE-based approaches typically exhibit more stable training behavior and clearer probabilistic semantics, but often face a trade-off when generating fine-grained trajectories.

In recent years, diffusion models have been introduced to trajectory generation.
They leverage iterative noise injection and denoising processes to synthesize high-quality trajectories.
Compared to GAN-based and VAE-based approaches, diffusion models typically offer more stable training dynamics and demonstrate strong performance, making them an increasingly popular paradigm for trajectory generation.
DiffTraj~\cite{zhu2023difftraj} applies diffusion probabilistic models to GPS trajectory generation and adopts a multi-scale feature fusion denoising network to preserve spatiotemporal structures. 
\cite{chu2024simulating} further incorporates road topology, enabling generated trajectories to satisfy road network constraints. 
ControlTraj~\cite{zhu2024controltraj} introduces explicit conditioning on road structures and trip attributes within the diffusion process for controllable trajectory generation.
To further improve feasibility within road networks, Diff-RNTraj~\cite{wei2024diff} represents trajectories as road segments with relative  proportions and introduces a spatial reachability loss to enforce network-level reachability. \cite{song2024controllable} incorporates user profile information during denoising, enabling personalized trajectory generation. 
SEED~\cite{rao2025seed} explores the integration of sequence modeling techniques with diffusion models to improve modeling efficiency while maintaining high generation quality.

With the rapid development of large language models (LLMs), trajectory generation approaches built on LLMs have attracted increasing attention. 
The core challenge of these methods lies in tokenizing the spatial and temporal attributes of trajectories into sequential representations that can be processed by large language models, and leveraging the strong sequence modeling and contextual reasoning capabilities of LLMs to generate  trajectories.
\cite{kobayashi2023modeling} uses a hierarchical spatial encoding scheme that maps location and temporal information into discrete tokens, and performs autoregressive trajectory generation using GPT-2. 
\cite{wang2024large} treats LLMs as decision-making agents, generating personalized mobility trajectories through activity pattern recognition and travel motivation reasoning.
\cite{shao2024beyond} exploits the reasoning abilities of LLMs by formulating trajectory generation as a context-aware commonsense reasoning process.

Despite the substantial progress achieved by all these trajectory generation methods, most of these approaches rely on a common implicit assumption that at least a subset of real mobility trajectories from the target city is accessible during model training or model construction.
In many real-world scenarios, however, this assumption is difficult to satisfy. Besides commercial barriers and privacy concerns, the high costs associated with sensor deployment, as well as the long-term burden of data management, prevent many cities from collecting and maintaining large-scale, high-resolution mobility trajectory datasets. 
As a result, the applicability of these existing trajectory generation methods is substantially limited in such data-inaccessible target cities.

\begin{figure}
    \centering
    \centerline{\includegraphics[scale=0.15]{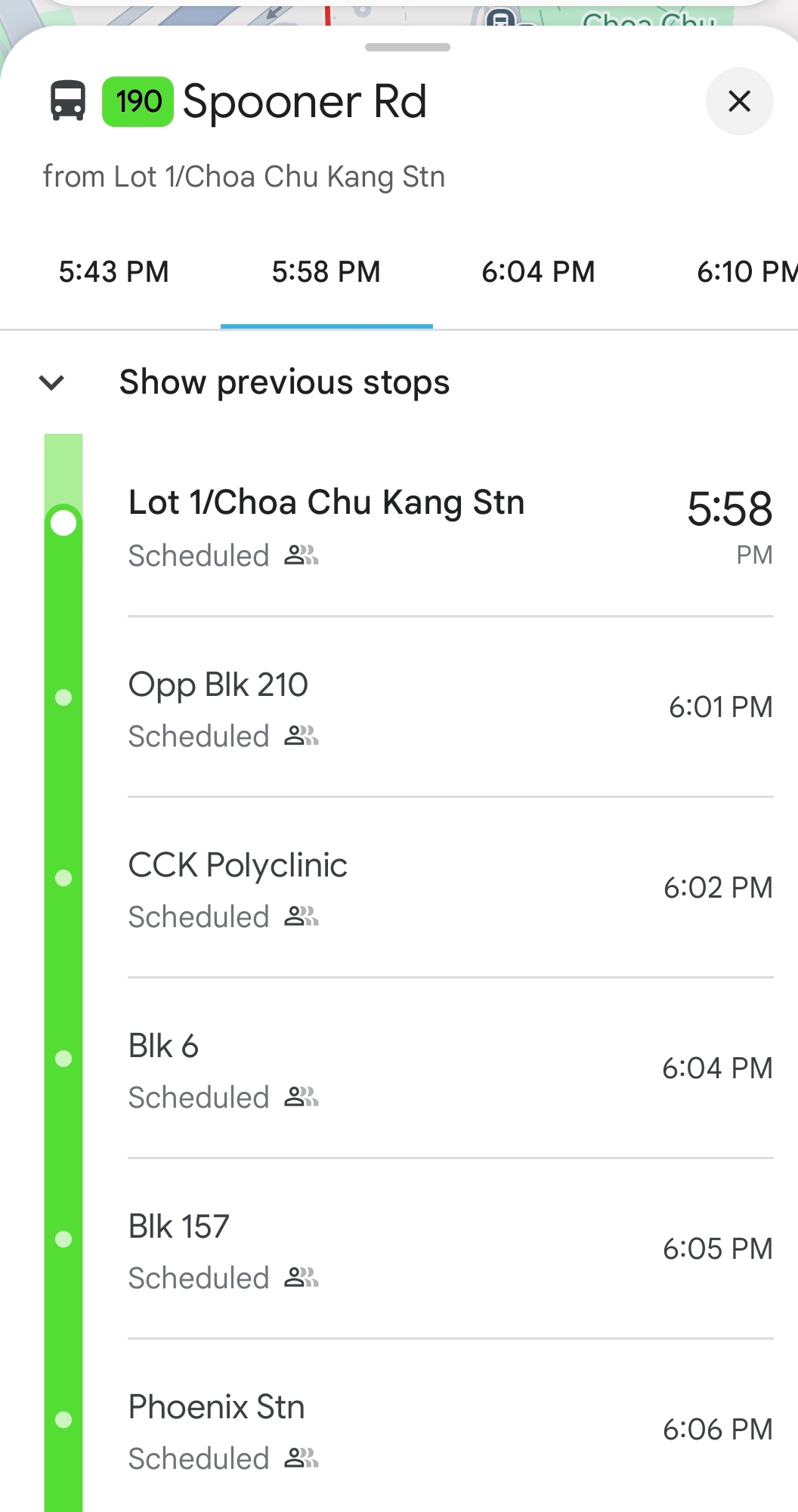}}
    \caption{
        Screenshot of bus  timetable in Google Map APP.
    }
    \label{fig:google map}
\end{figure}

\subsection{Task Arithmetic}
\label{sec:detail ta}
Task arithmetic has recently been a powerful approach for efficient model editing and generalization. 
It is based on task vectors, which are defined as the difference between parameters of a fine-tuned model and parameters of its corresponding pre-trained model. 
Task vectors serve as compact representations of specific tasks in parameter space, enabling direct and efficient modifications to a model’s behavior without full retraining~\cite{ilharco2022editing,zhang2023composing}.
Their effectiveness has been supported by several works.
\cite{zhou2024metagpt} conducts a detailed analysis of loss differences under vector operations. 
They derive upper bounds for both individual and average task loss differences, and further provide a closed-form solution for the optimal scaling coefficient that minimizes the average loss discrepancy between composed tasks. 
\cite{li2025task} provides theoretical validation of task arithmetic in nonlinear Transformer architectures. 
They prove that, when tasks are independent, vector addition enables successful multi-task generalization.
They also show that subtractive composition allows for precise forgetting in the presence of conflicting tasks. 
Moreover, this paper proves that, by linearly combining vectors from a set of unrelated tasks with appropriate coefficients,  a model that generalizes to entirely new tasks can be constructed.

Beyond theoretical works, task arithmetic has proven to be a practical tool for both injecting new capabilities and eliminating harmful behaviors.
\cite{huang2024chat} demonstrates how instruction-following capabilities learned via fine-tuning can be transferred to a base model through vector addition.
TIES-Merging~\cite{yadav2023ties}  resolves parameter sign conflicts and pruning insignificant updates, enabling multi-task abilities to be merged without performance loss.
In behavior removal and safety alignment, task arithmetic offers a lightweight yet effective solution.
\cite{bhardwaj2024language} shows that safety-related behavior lost during fine-tuning can be restored by directly adding a “safety vector.” 
Ethos~\cite{gao2024ethos} introduces an orthogonal subspace projection method, decomposing task vectors into general-purpose and harmful components. 
By removing  projection along harmful directions, Ethos enables targeted suppression of toxicity and bias, achieving precise and interpretable model correction.

Task arithmetic has also demonstrated strong potential in addressing out-of-distribution tasks and enabling cross-domain generalization, especially in low-resource or zero-shot settings.
~\cite{parovic2024investigating} decouples language vectors and task vectors, and recombines them to support zero-shot transfer across languages.
\cite{chronopoulou2024language} leverages compositional task vectors to enable summarization in languages with no labeled data.
In domain-specific settings, task arithmetic helps bridge gaps between synthetic and real-world data. 
In information retrieval, domain-specific vectors significantly enhance model performance in unseen fields such as medicine and science~\cite{braga2025investigating}. 
In automatic speech recognition, task vector differences between synthetic and real data guide the model toward realistic behavior distributions, improving performance in real-world environments~\cite{su2024task}.

\section{Screenshot of Bus Timetable Provided by Different Map Apps}
\begin{figure}
\centering
    \includegraphics[scale=0.15]{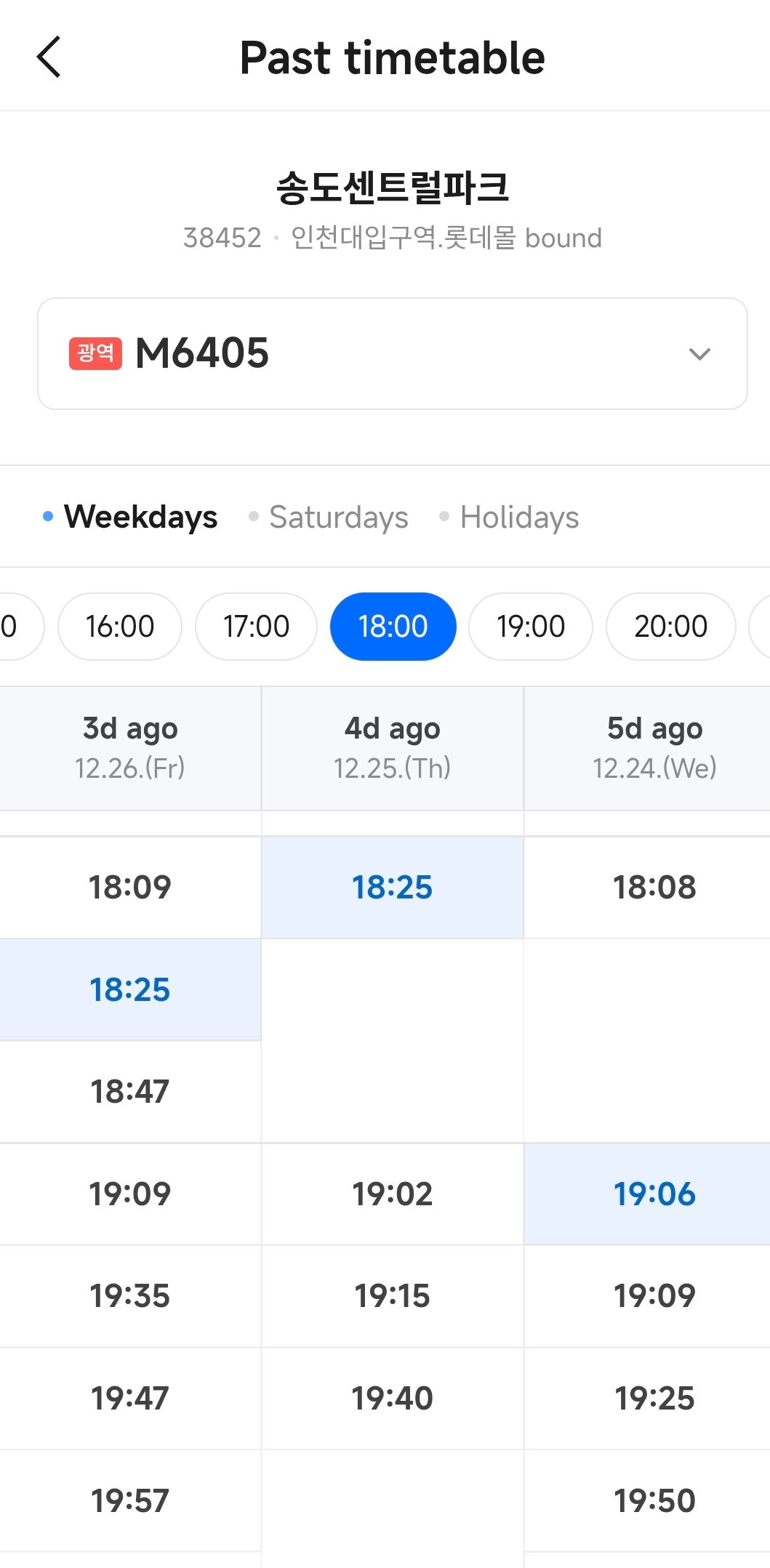}
    \caption{
        Screenshot of bus  timetable in Naver Map APP.
    }
    \label{fig:Naver}
    \vspace{-15pt}
\end{figure}

Fig.~\ref{fig:google map} and \ref{fig:Naver} show the screenshot of bus timetable provided by Google Map and Naver Map. The former is widely used in numerous countries and the latter is used in Korea.

\section{Detailed Experimental Details of Empirical Observations on Bus Timetables and Mobility Trajectories}

\subsection{Detailed Experimental Setup for Spatial Stability of Bus Station Sequences}
\label{sec:detail spatial sta}
Here we provide a detailed explanation of the experiment which analyze the spatial consistency of bus station sequences in Shanghai between 2015 and 2025.

\subsubsection{Data and Preprocessing}

The  bus station sequence data in 2015 is collected from a public historical dataset \footnote{\url{https://msittig.wubi.org/bus/}}, which retains at most the first 30 bus stops per route. 
In contrast, the 2025 bus station sequence data is from its corresponding bus timetable data, which is crawled from the Moovit platform~\footnote{\url{https://moovitapp.com/}}.
For consistency, the 2025 bus station sequence data is also truncated to the first 30 bus stops per route, ensuring that route comparisons across years are performed on sequences of equal maximum length.
To reduce noise caused by inconsistencies in bus station naming, we apply lightweight normalization to all bus station names, including the removal of whitespace and any parenthetical annotations. 
These normalized bus station names are used for all similarity computations in this subsection.
To enable alignment across years, we group bus routes by their route IDs, with routes sharing the same ID treated as belonging to the same group.
Since the same route ID may correspond to different travel directions or terminal stations across years, a one-to-one correspondence between 2015 and 2025 bus lines does not always exist. 
Within each route group, we enumerate all feasible route pairings and select the pairing with the highest station-sequence similarity.
Across the ten-year span from 2015 to 2025, approximately 88\% of bus routes in the 2025 dataset can be successfully matched to a corresponding route group from 2015.
Similarity results are then computed based on all successfully matched route pairs, reflecting the spatial consistency of the majority of bus routes that persist across the ten-year span.
After route matching, we separately compute summary statistics for (a) routes within the entire city, and (b) routes within Outer Ring Road, which is our study region for Shanghai.

\subsubsection{Similarity Metrics for Evaluating Spatial Stability of Bus Station Sequences}

Given the full set of bus station sequences in 2015:
\begin{equation}
\begin{aligned}
R_{2015}^{\text{bus}}
&= \left\{ r_{2015,1}^{\text{bus}}, r_{2015,2}^{\text{bus}}, \dots, r_{2015,N_{2015}}^{\text{bus}} \right\}, \\
r_{2015,i}^{\text{bus}}
&= \left( s_{i,1}^{2015}, s_{i,2}^{2015}, \dots, s_{i,n_i}^{2015} \right)
\end{aligned}
\end{equation}
and  the  full set of  bus station sequences in 2025:
\begin{equation}
\begin{aligned}
R_{2025}^{\text{bus}}
&= \left\{ r_{2025,1}^{\text{bus}}, r_{2025,2}^{\text{bus}}, \dots, r_{2025,N_{2025}}^{\text{bus}} \right\}, \\
r_{2025,j}^{\text{bus}}
&= \left( s_{j,1}^{2025}, s_{j,2}^{2025}, \dots, s_{j,m_j}^{2025} \right).
\end{aligned}
\end{equation}

we compute  similarity metrics for each matched pair $\left( r_{2015,i}^{\text{bus}}, r_{2025,j}^{\text{bus}} \right)$.
Final statistics are aggregated over all such matched pairs' results.

Longest Common Subsequence (LCS) measures the maximum length of a subsequence that is common to two ordered bus station sequences while preserving relative order.
Formally, let $||LCS(x, y)||$ denote the length of the longest common subsequence between sequences x and y. 
The normalized LCS similarity between a matched pair of bus station sequences $r^{bus}_{2015,i}$ and $r^{bus}_{2025,j}$ is defined as
Its normalized form is:
\begin{equation}
    \text{LCS}_{\text{ratio}}(i,j) = \frac{||\text{LCS}(r_{2015,i}^{\text{bus}}, r_{2025,j}^{\text{bus}})||}{\min(n_i, m_j)},
\end{equation}
where $n_i$ and $m_j$ are the number of bus stops in the respective bus station sequence.
This metric is sensitive to order but robust to insertions and deletions.

Jaccard Similarity (Jac) measures the  overlap of bus stops between two bus station sequences.
Let $S(r) = \left\{ s \mid s \in r \right\}$ be the set of bus stops of bus station sequence $r$, we have the following definition of Jaccard Similarity:
\begin{equation}
    \text{Jaccard}(i,j) = \frac{ \left| S\left( r_{2015,i}^{\text{bus}} \right) \cap S\left( r_{2025,j}^{\text{bus}} \right) \right| }{ \left| S\left( r_{2015,i}^{\text{bus}} \right) \cup S\left( r_{2025,j}^{\text{bus}} \right) \right| }
\end{equation}

Dice Similarity is another set-based metric, which is more sensitive to the size of the intersection. 
It is defined as
\begin{equation}
    \mathrm{Dice}(i,j) =
\frac{2 \left| S(r^{\text{bus}}_{2015,i}) \cap S(r^{\text{bus}}_{2025,j}) \right|}
{\left| S(r^{\text{bus}}_{2015,i}) \right| + \left| S(r^{\text{bus}}_{2025,j}) \right|}.
\end{equation}

Edit Distance on Real sequences (EDR) quantifies the minimum number of edit operations, including insertions, deletions, or substitutions, required to transform one bus station sequence into another. 
Smaller values indicate greater structural similarity.

Dynamic Time Warping (DTW) aligns sequences of unequal length by minimizing cumulative matching cost.
In our experiment, the matching cost is defined based on whether the name of bus stops are the same.
Lower DTW values indicate closer alignment in bus station sequences.

\subsubsection{Detailed Analysis and Discussion}

Table~\ref{sh bus 2015} quantitatively reports the spatial similarity between Shanghai’s bus station sequences in 2015 and 2025.
All similarity metrics consistently yield high scores, indicating that the spatial structure of bus routes remains largely stable over a ten-year time span.

For order-sensitive metrics,  LCS  reaches 0.8109 at the citywide level, implying that more than 80\% of bus stops can be aligned while preserving relative order for most bus routes. 
This suggests that although some bus routes may undergo stop adjustments or branch modifications, their overall spatial structure remains stable. 
Consistently,  EDR and DTW values are 0.2522 and 0.1355, respectively, indicating relatively low edit and alignment costs when transforming 2015 bus sequences into their 2025 counterparts.
From a set-level perspective,  Jaccard and Dice similarities achieve  0.6975 and 0.7937, respectively, demonstrating substantial overlap in bus stop coverage between the two set of bus station sequences. 
The higher Dice score, which emphasizes intersection size, further indicates that the majority of bus stops are retained over time.

When such analysis is restricted to area within the Outer Ring Road, which is the central area of shanghai, all metrics show further improvement.
LCS increases to 0.8546, indicating greater order stability in the urban area. Jaccard and Dice similarities rise to 0.7459 and 0.8383, respectively, while EDR and DTW decrease to 0.2011 and 0.1062. 
These results suggest that structural adjustments to bus routes are even more limited in the urban area, and spatial  stability is more guaranteed there.

\subsection{Detailed Experimental Setup for Spatiotemporal Correlations Between Bus Timetables and Mobility Trajectories}
\label{sec:bus timetable relationship detail}
To analyze the spatiotemporal correlations between bus timetables and general mobility trajectories, we partition the study area of each city into regular grids with $1km \times 1km$ resolution.
Subsequently, all bus station locations and mobility trajectory points are mapped to their corresponding grid cells.

\subsubsection{Spatial Distribution Consistency Metrics}

Let $c_i^{\text{bus}}$ and $c_i^{\text{{mob}}}$ be the counts of bus stops from bus timetables and mobility trajectory points that fall into grid cell $i$, respectively.
We first define the normalized spatial distributions as
\begin{equation}
    P_i^{\text{bus}} = \frac{c_i^{\text{bus}}}{\sum_j c_j^{\text{bus}}}, \qquad
P_i^{\text{mob}} = \frac{c_i^{\text{mob}}}{\sum_j c_j^{\text{mob}}}.
\end{equation}

To systematically evaluate the spatial alignment between bus timetables and mobility trajectories, we adopt multiple complementary metrics, including Jensen–Shannon Divergence ($\mathrm{JSD}_c$), Cosine Similarity (Cos), and Spatial Pearson Correlation (Pear-S).
Jensen–Shannon Divergence (JSD) is used to quantify the overall discrepancy between the two distributions:
\begin{equation}
    \mathrm{JSD}_c\!\left(P^{\text{bus}} \,\|\, P^{\text{mob}}\right)
= \frac{1}{2}\mathrm{KL}\!\left(P^{\text{bus}} \,\|\, M\right)
+ \frac{1}{2}\mathrm{KL}\!\left(P^{\text{mob}} \,\|\, M\right),
\end{equation}
where
\begin{equation}
    M = \frac{1}{2}\left(P^{\text{bus}} + P^{\text{mob}}\right).
\end{equation}
Cosine similarity (Cos) is computed to measure directional difference between the two spatial distributions:
{\fontsize{8.5pt}{8pt}\selectfont
\begin{equation}
    \mathrm{Pear\text{-}S}(P^{bus},P^{{Mob}})
= \frac{\sum_i \left(P_i^{\text{bus}} - \bar{P}^{\text{bus}}\right)
\left(P_i^{\text{mob}} - \bar{P}^{\text{mob}}\right)}
{\sqrt{\sum_i \left(P_i^{\text{bus}} - \bar{P}^{\text{bus}}\right)^2}
\sqrt{\sum_i \left(P_i^{\text{mob}} - \bar{P}^{\text{mob}}\right)^2}}
\end{equation}
}

\subsubsection{Spatial Coverage Metrics}
To further analyze the spatial coverage capacity of bus timetables, we construct urban travel corridors based on bus timetables. Specifically, for a bus timetable $\tau^{\text{bus}} = \{(s_1, \tilde{t}_1), (s_2, \tilde{t}_2), \ldots, (s_m, \tilde{t}_m)\}$, we represent its spatial trajectory as a polyline, denoted as
\begin{equation}
    \ell^{\text{bus}} = \mathrm{LineString}(s_1, s_2, \ldots, s_m).
\end{equation}
We apply a buffer with fixed radius $r$ to each bus polyline $\ell_k^{\text{bus}}$ to obtain a bus corridor region $\mathrm{Buffer}\!\left(\ell_k^{\text{bus}}, r\right)$. 
In all experiments, we set the buffer radius to $r=100m$, which reflects a typical pedestrian access distance to bus routes in urban environments~\cite{daniels2013explaining,tetali2025developing}.
Aggregating all such buffered corridors yields the city-scale bus corridor region:
\begin{equation}
    \mathcal{C}^{\text{bus}} = \bigcup_k \mathrm{Buffer}\!\left(\ell_k^{\text{bus}}, r\right).
\end{equation}
Point Coverage is used to quantify the proportion of mobility trajectory points that fall within bus corridors. 
Let $\{p_i\}$ denote all points from mobility trajectories. 
The point coverage is defined as
\begin{equation}
    \mathrm{PCov}
= \frac{1}{|\{p_j\}|} \sum_j \mathbf{1}\!\left(p_j \in \mathcal{C}^{\text{bus}}\right),
\end{equation}
where $\mathbf{1}(\cdot)$ is the indicator function. 
Point Coverage reflects the extent to which mobility trajectories spatially overlap with bus transit structures at the point level.
To further capture path-level structural overlap, we define Route Coverage. 
Given a mobility trajectory represented as a polyline $\ell^{\text{mob}}$, its route coverage is defined as
\begin{equation}
    \mathrm{RCov_{traj}}(\ell^{\text{mob}})
= \frac{\mathrm{Length}\!\left(\ell^{\text{mob}} \cap \mathcal{C}^{\text{bus}}\right)}
{\mathrm{Length}\!\left(\ell^{\text{mob}}\right)}.
\end{equation}
Averaging over all mobility trajectories yields the overall route coverage:
\begin{equation}
    \mathrm{RCov}
= \mathbb{E}_{\ell^{\text{mob}}}
\left[\mathrm{RCov_{traj}}(\ell^{\text{mob}})\right].
\end{equation}
Point coverage reflects how frequently mobility trajectories are near bus corridors, while route coverage characterizes the extent to which mobility trajectories extend along bus station sequences. 
Together, these metrics provide complementary perspectives on the spatial representational capacity of bus transit trajectories for general urban mobility.

\subsubsection{Temporal Correlation Metrics}

Beyond spatial correlations, we further examine the temporal correlation between bus timetables and mobility trajectories. 
Each day is divided into 24 hourly intervals, and the  frequencies of bus timetables and mobility trajectories are computed for each interval. 
The normalized temporal distributions are denoted as
\begin{equation}
\mathbf{T}^{\text{bus}} = (T_1^{\text{bus}}, \ldots, T_{24}^{\text{bus}}), \qquad
\mathbf{T}^{\text{mob}} = (T_1^{\text{mob}}, \ldots, T_{24}^{\text{mob}}),
\end{equation}
where
\begin{equation}
    \sum_{h} T_h^{\text{bus}} = \sum_{h} T_h^{\text{mob}} = 1.
\end{equation}
Temporal correlation is computed using the temporal Pearson correlation coefficient:
{\fontsize{8.5pt}{8pt}\selectfont
\begin{equation}
    \mathrm{Pear\text{-}T}(\mathbf{T}^{\text{bus}},\mathbf{T}^{\text{mob}})
    = \frac{\sum_h \left(T_h^{\text{bus}} - \bar{\mathbf{T}}^{\text{bus}}\right)
    \left(T_h^{\text{mob}} - \bar{\mathbf{T}}^{\text{mob}}\right)}
    {\sqrt{\sum_h \left(T_h^{\text{bus}} - \bar{\mathbf{T}}^{\text{bus}}\right)^2}
    \sqrt{\sum_h \left(T_h^{\text{mob}} - \bar{\mathbf{T}}^{\text{mob}}\right)^2}}
\end{equation}
}
where $\bar{\mathbf{T}}^{\text{bus}}$ and $\bar{\mathbf{T}}^{\text{mob}}$ denote the mean values of the corresponding temporal distributions.

\begin{figure}
  \centering
  \includegraphics[scale=0.35]{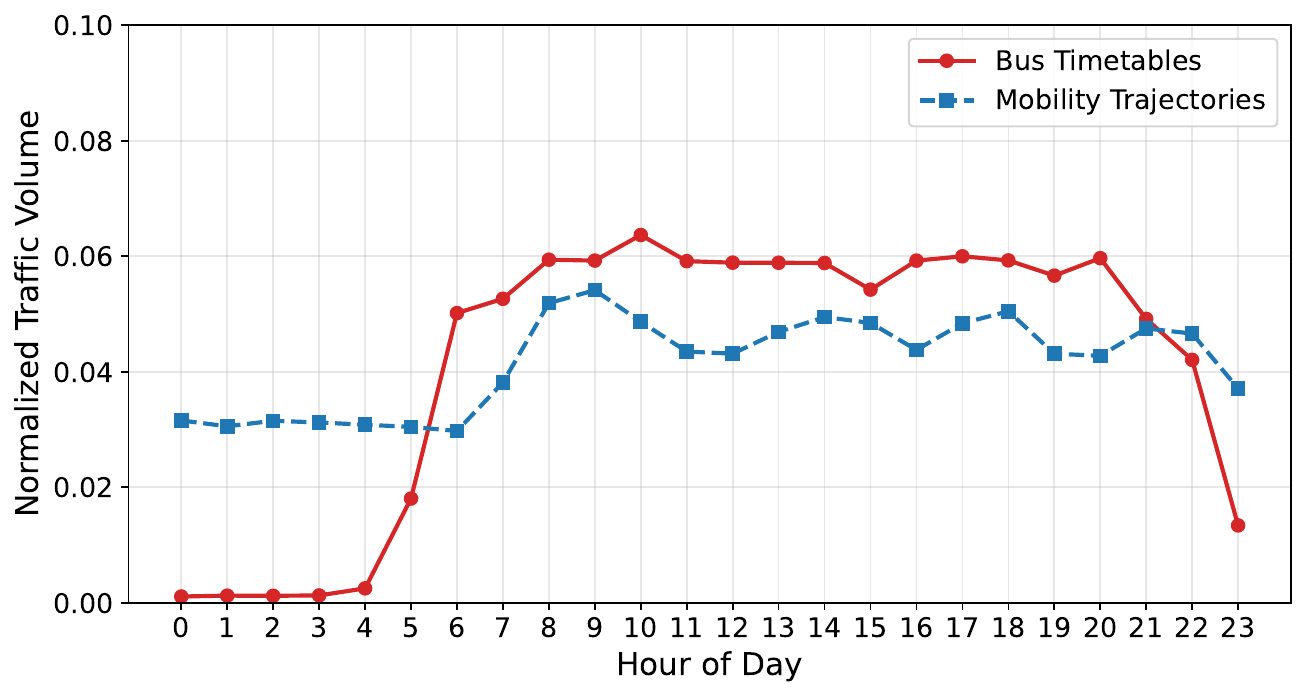}
    \caption{
        Hourly temporal distribution of bus timetables and mobility trajectories in Shanghai.
    }
    \label{fig:sh tem dis}
\end{figure}

\begin{figure}
  \centering
  \includegraphics[scale=0.35]{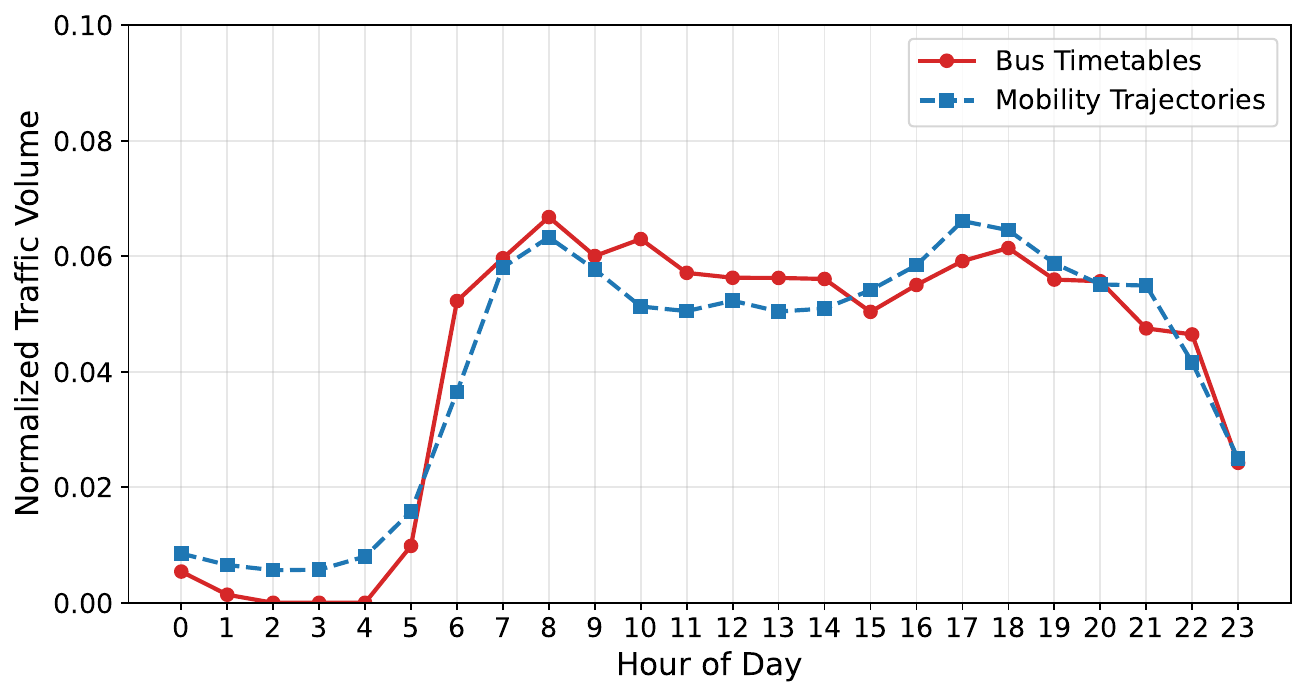}
    \caption{
        Hourly temporal distribution of bus timetables and mobility trajectories in Singapore.
    }
    \label{fig:sg tem dis}
\end{figure}

\begin{figure}
  \centering
  \includegraphics[scale=0.35]{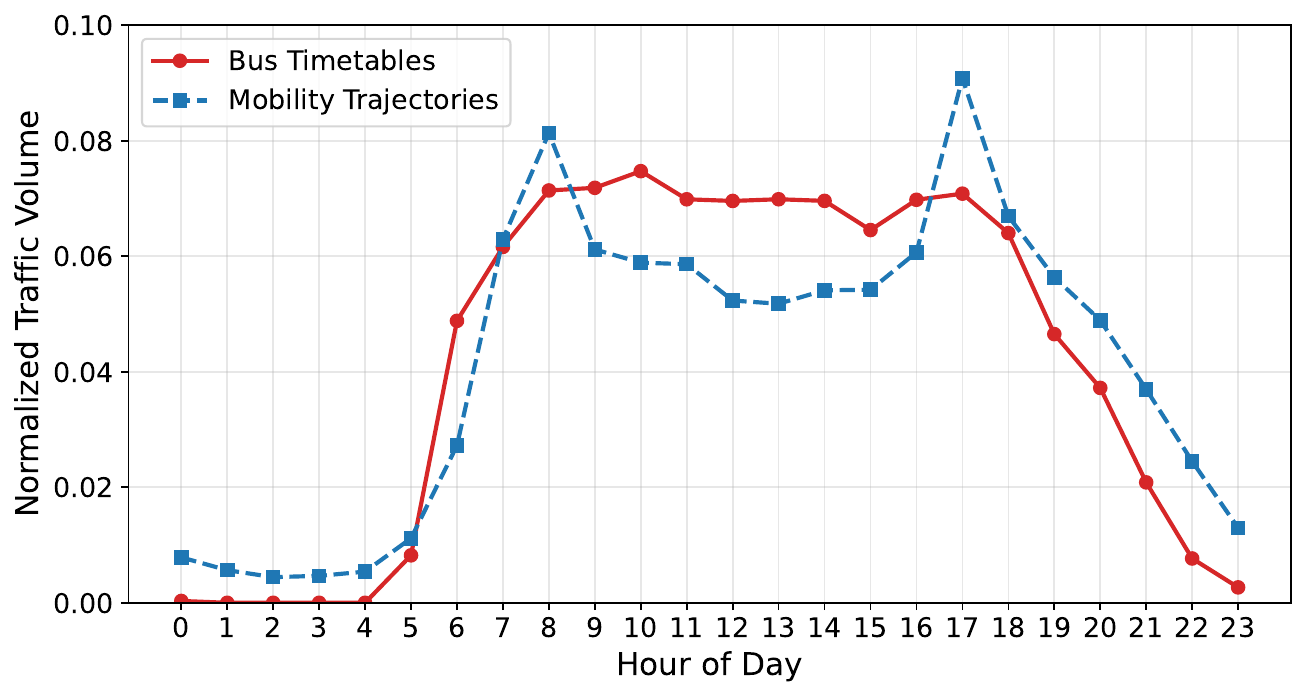}
    \caption{
        Hourly temporal distribution of bus timetables and mobility trajectories in Wuxi.
    }
    \label{fig:wx tem dis}
\end{figure}

\section{Spatial Discretization}
\label{sec:spatial app}
\begin{figure*}
  \vskip 0.2in
  \begin{center}
    \centerline{\includegraphics[scale=0.35]{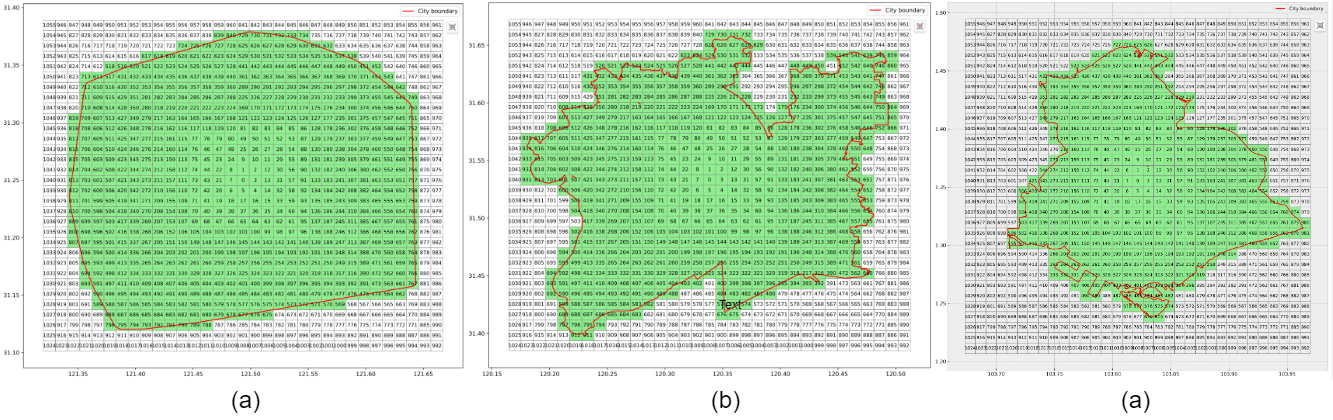}}
    \caption{
         Visualization of spatial discretization and the spiral indexing strategy across different cities.  Grid labeled with  index $i$  corresponds to the spatial token \texttt{<LOC\_i>}.
    }
    \label{fig:grid}
  \end{center}
\end{figure*}

To ensure structural comparability across different cities, we employ a unified spiral indexing strategy for spatial discretization. 
Figure~\ref{fig:grid} provides visualizations of this strategy applied to Shanghai, Wuxi, and Singapore. 
In each representation, the study area is partitioned into regular grids with the same spatial resolution ($1km \times 1km$). 
The grid located at the geometric center of the study area, which is computed from the boundary polygon of urban center area, is designated as the anchor location and assigned the index $0$ (corresponding to the token \texttt{<LOC\_0>}). 
Subsequent indices are assigned to surrounding grids in an outward, clockwise spiral manner. 
 Formally,  grid labeled with  index $i$  corresponds to the spatial token \texttt{<LOC\_i>}.
Grids that fall within the predefined city boundary are highlighted in green, while those outside are shown in white. 
This consistent indexing scheme allows relative spatial positions (e.g., center vs. periphery) to be encoded similarly across different cities, facilitating unified trajectory representation.

\section{First-order Approximation for Projected Task Vector Difference in task-relevant Parameter Subspace}
\label{sec:detail theory}
This section provides a  detailed  derivation of the key claims made in Sec.~\ref{first order}.
Specifically, we show that for a trajectory dataset $D$, the projection of the task vector difference between base and instruction-tuned LLMs onto the task-relevant parameter subspace $\mathcal{T}_{\mathrm{traj}}$ satisfies
\begin{equation}
\begin{aligned}
&\mathbf{P}_{\mathrm{traj}}\bigl(\mathbf{V}_B(\mathcal{D}) - \mathbf{V}_I(\mathcal{D})\bigr) \approx \\
&\quad -\eta \,\mathbf{P}_{\mathrm{traj}}\bigl(\nabla_{\theta} \mathcal{L}(\theta_0^{(B)}; \mathcal{D}) - \nabla_{\theta} \mathcal{L}(\theta_0^{(I)}; \mathcal{D})\bigr)
\end{aligned}
\end{equation}
We further demonstrate that, under a first-order approximation, this projected difference is expected to remain numerically small.

\subsection{First-Order Approximation of Task Vectors}

Let $\theta$ denote the model parameters, and let  $\mathcal{L}(\theta; \mathcal{D})$ denote the trajectory generation training loss on trajectory dataset $\mathcal{D}$.
Let $\theta_0$ denote the initialization parameters for fine-tuning.
A parameter-efficient fine-tuning  procedure involving $T$ gradient descent steps produces a parameter sequence $\{\theta_t\}_{t=0}^{T}$ via the update rule:
\begin{equation}
\theta_{t+1}
=
\theta_t
-
\eta_t \nabla_{\theta} \mathcal{L}(\theta_t; \mathcal{D}),
\quad
t = 0, \dots, T-1,
\end{equation}
where $\eta_t$ is the learning rate at step $t$.
The total parameter shift, defined as the task vector $\mathbf{V}(\theta_0,\mathcal{D})$, is therefore given by:
\begin{equation}
\mathbf{V}(\theta_0,\mathcal{D}):=\theta_T - \theta_0
=
-\sum_{t=0}^{T-1} \eta_t \nabla_{\theta} \mathcal{L}(\theta_t; \mathcal{D}),
\end{equation}
Under a small-step and parameter-efficient fine-tuning setting, parameter updates are small, which
suppresses higher-order terms and makes a first-order approximation a
reasonable abstraction.
Thus, the optimization dynamics can be analyzed via a first-order Taylor expansion around the initialization $\theta_0$. 
This yields the standard linearized task vector approximation:
\begin{equation}
\mathbf{V}(\theta_0,\mathcal{D})
\approx
-\eta \nabla_{\theta} L(\theta_0; D).
\end{equation}
where $\eta$ represents the effective step size.
Applying this approximation to the base and instruction-tuned LLMs' initialization parameters $\theta_0^{(B)}$ and $\theta_0^{(I)}$, respectively, we will get:
\begin{equation}
\mathbf{V}_B(\mathcal{D}) - \mathbf{V}_I(\mathcal{D})
\approx
-\eta
\bigl(
\nabla_{\theta} \mathcal{L}(\theta_0^{(B)}; \mathcal{D})
-
\nabla_{\theta} \mathcal{L}(\theta_0^{(I)}; \mathcal{D})
\bigr).
\end{equation}
Since the projection operator $\mathbf{P}_{\mathrm{traj}}$ is linear, applying it to both sides  gives Eq.28.

\subsection{ First-Order Weak Coupling of the Projected Gradient Shift}
We introduce the projected gradient function
\begin{equation}
g_T(\theta; \mathcal{D})
:=
\mathbf{P}_{\mathrm{traj}} \nabla_{\theta} \mathcal{L}(\theta; \mathcal{D}),
\end{equation}
where $\mathbf{P}_{\mathrm{traj}}$ denotes the orthogonal projection onto the task-relevant parameter subspace $\mathcal{T}_{\mathrm{traj}}$.
The projection operator $\mathbf{P}_{\mathrm{traj}}$ is defined with respect to the task-relevant subspace
$\mathcal{T}_{\mathrm{traj}}$ induced at the reference initialization
$\theta_0^{(B)}$ and is treated as a constant linear operator
independent of \(\theta\).
Under this convention, $\mathbf{P}_{\mathrm{traj}}$  is a fixed linear operator, and it
follows that
\begin{equation}
\nabla_{\theta} g_T(\theta; \mathcal{D})
=
\mathbf{P}_{\mathrm{traj}} \nabla_{\theta}^2 \mathcal{L}(\theta; \mathcal{D})
=
\mathbf{P}_{\mathrm{traj}} H(\theta; \mathcal{D}),
\end{equation}
where $H(\theta; \mathcal{D})$ denotes the Hessian matrix of trajectory generation loss  $\mathcal{L}(\theta; \mathcal{D})$ with respect to the parameters $\theta$.

We perform a first-order Taylor expansion of $g_T(\theta; \mathcal{D})$ around the
reference initialization $\theta_0^{(B)}$ and substitute $\theta_0^{(I)} = \theta_0^{(B)} + \delta_0$, obtaining
\begin{equation}
g_T(\theta_0^{(I)}; \mathcal{D})
=
g_T(\theta_0^{(B)}; \mathcal{D})
+
\mathbf{P}_{\mathrm{traj}} H(\theta_0^{(B)}; \mathcal{D})\, \delta_0
+
O(\|\delta_0\|^2).
\end{equation}
\
Rearranging terms gives
\begin{equation}
g_T(\theta_0^{(B)}; \mathcal{D})
-
g_T(\theta_0^{(I)}; \mathcal{D})
=
-
\mathbf{P}_{\mathrm{traj}} H(\theta_0^{(B)}; \mathcal{D})\, \delta_0
+
O(\|\delta_0\|^2).
\end{equation}
Let  $\mathbf{Q}_{\mathrm{traj}} := \mathbf{I} - \mathbf{P}_{\mathrm{traj}}$.
Any vector, including $\delta_0$, admits the decomposition
\begin{equation}
\delta_0 = \mathbf{P}_{\mathrm{traj}}\delta_0 + \mathbf{Q}_{\mathrm{traj}}\delta_0,
\quad
\mathbf{P}_{\mathrm{traj}}\delta_0 \in \mathcal{T}_{\mathrm{traj}},
\;
\mathbf{Q}_{\mathrm{traj}}\delta_0 \in \mathcal{T}_{\mathrm{traj}}^{\perp}.
\end{equation}

Substituting this decomposition into $\mathbf{P}_{\mathrm{traj}} H(\theta_0^{(B)}; \mathcal{D})\, \delta_0$ yields
\begin{equation}
\begin{aligned}
    \mathbf{P}_{\mathrm{traj}} H(\theta_0^{(B)}; \mathcal{D}) \delta_0 
    &= \mathbf{P}_{\mathrm{traj}} H(\theta_0^{(B)}; \mathcal{D}) \mathbf{P}_{\mathrm{traj}} \delta_0 \\
    &\quad + \mathbf{P}_{\mathrm{traj}} H(\theta_0^{(B)}; \mathcal{D}) \mathbf{Q}_{\mathrm{traj}} \delta_0.
\end{aligned}
\end{equation}
The first term on the right-hand side is governed by the projection of the initialization difference $\delta_0$ onto the task-relevant parameter subspace $\mathcal{T}_{\mathrm{traj}}$.
$\delta_0$ is primarily associated with instruction-following and safety alignment rather than trajectory-related tokens, and it lies predominantly in the orthogonal complement $\mathcal{T}_{\mathrm{traj}}^\perp$, rendering the projection $\|\mathbf{P}_{\mathrm{traj}}\delta_0\|$ small even when $||\delta_0||$ itself might not. 
The second term on the right-hand side captures the interaction between $\mathcal{T}_{\mathrm{traj}}$ and its orthogonal complement $\mathcal{T}_{\mathrm{traj}}^\perp$.
Under the setting of MobTA, given that the trajectory generation dynamics are structurally disentangled from specific instruction-following and safety alignment tasks, parameter perturbations in $\mathcal{T}_{\mathrm{traj}}^\perp$ induce negligible gradient changes along trajectory-relevant directions, making the second term on the right-hand side small as well.
Consequently, the first-order contribution of  $\mathbf{P}_{\mathrm{traj}} H(\theta_0^{(B)}; \mathcal{D})\, \delta_0$ is suppressed and the dominant contribution to $||g_T(\theta_0^{(B)}; \mathcal{D})
-
g_T(\theta_0^{(I)}; \mathcal{D})||$ arises from higher-order terms of order $O(\|\delta_0\|^2)$,  which are themselves minimal under a small-step and parameter-efficient fine-tuning setting~\cite{jacot2018neural,hu2022lora,malladi2023kernel}, making $||\mathbf{P}_{\mathrm{traj}} H(\theta_0^{(B)}; \mathcal{D})\, \delta_0||$ numerically negligible.
This leads to $||g_T(\theta_0^{(B)}; \mathcal{D})
-
g_T(\theta_0^{(I)}; \mathcal{D})||$, and further $||\mathbf{P}_{\mathrm{traj}}\bigl(\mathbf{V}_B(\mathcal{D}) - \mathbf{V}_I(\mathcal{D})\bigr)||$, numerically negligible.

\section{Data and Data Processing}
\label{app data}
\subsection{Bus Timetable Collection and Station Coordinate Extraction}
\begin{figure*}
  \vskip 0.2in
  \begin{center}
    \centerline{\includegraphics[scale=0.5]{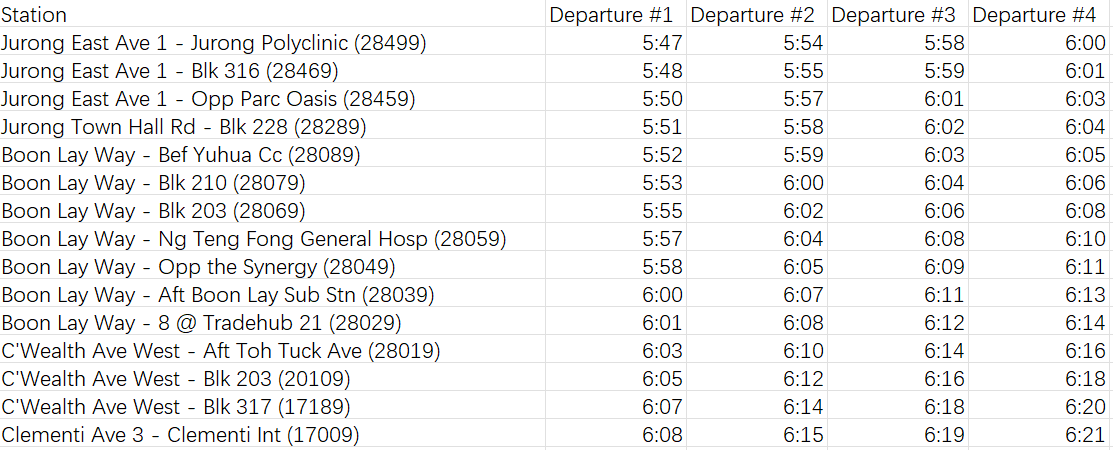}}
    \caption{
        Example of one bus transit timetable in Singapore. Bus transit timetables in Shanghai and Wuxi are in the same format.
    }
    \label{fig:example}
  \end{center}
\end{figure*}

We collect bus timetables from Moovit platform~\footnote{\url{https://moovitapp.com/}}.
Specifically, for each city, we crawl all available bus route pages, parse the complete bus station sequences and scheduled arrival times for each route, and convert all extracted data into a unified “station × departure time” tabular format. 
This process yields standardized station-level bus  timetable data. 
Example of one processed Singapore bus timetable is illustrated in Fig.\ref{fig:example}.

After obtaining  bus timetables for each city, we further augment each bus stop with corresponding geographic coordinates. 
For Singapore, bus stop latitude and longitude information is obtained from the publicly released static dataset of the Land Transport Authority (LTA)~\footnote{\url{https://datamall.lta.gov.sg/content/datamall/en/static-data.html}}. 
We match the bus stop IDs extracted from Moovit with the official LTA records to assign  geographic coordinates to each bus stop.
For Shanghai and Wuxi, bus stop coordinates are queried via the Amap  API. 
Since some bus stops are missing from public map services or exhibit obvious spatial deviations in queried coordinates, we conduct careful manual inspection and validation for bus stations in these two cities to ensure spatial accuracy. 
Moreover, as Amap adopts the GCJ-02 coordinate system, all coordinates are converted into the standard WGS-84 coordinate system to maintain consistency across data sources.

\subsection{Spatiotemporal Cropping and Filtering }
For all three cities, our analysis focuses exclusively on their core urban areas. 
Both bus timetable data and mobility trajectory data are spatially cropped according to corresponding study boundaries, and only bus stations and trajectory records falling within these regions are retained. 
After spatial cropping, bus timetables and mobility trajectories with fewer than three records are discarded, as such short sequences lack meaningful spatiotemporal structure.
Along the temporal dimension, we discretize each day into minute-level intervals, resulting in a total of $24 \times 60=1440$ time steps. 
The resulting time values are subsequently represented using existing numerical tokens, rather than newly
introduced time-specific tokens.
Since time-of-day is inherently ordered, this
representation preserves relative temporal proximity and ordering information
through the numerical values themselves.
This unified temporal resolution ensures consistency across bus timetables and mobility trajectories in different cities.

\subsection{Grid-Based Representation and Trajectory Length Control}

To construct a unified spatial representation, all bus timetables and mobility trajectories are mapped to the regular $1km \times 1km$ grid cells in which they reside, thereby discretizing continuous geographic coordinates into grid-based location identifiers. 
Under this representation, both bus timetables and real mobility trajectories can be expressed as discrete sequences of time–location pairs.
Considering the computational and modeling constraints of large language models when handling long sequences, we apply uniform trajectory length control. 
For both bus timetables and real mobility trajectories, if the trajectory length exceeds 50, it is split into two shorter sub-trajectories. 
This operation is applied only to excessively long samples to prevent individual trajectories from occupying an excessive number of sequence tokens.
 Summary statistics of mobility trajectory data for each city are reported in Table~\ref{tab:dataset_statistics}.

\begin{table*}[t]
\centering
\caption{Dataset statistics.}
\label{tab:dataset_statistics}

\begin{tabular}{lccc}
\hline
Dataset & Num of Records & Num of Trajs & Time Span \\
\hline
SH & 18,462,492 & 1,007,133 & 2016-08-01 -- 2016-08-03 \\
WX      & 346,416    & 16,436    & 2020-07-18 -- 2020-08-17 \\
SG & 3,600,936  & 187,988   & 2024-03-01 -- 2024-03-31 \\
\hline
\end{tabular}

\end{table*}

\section{Baseline Introduction and Implementation Details}
\label{baseline detail}
Here we provide a brief introduction of baseline models we use, and implementation details of these baselines.
Except for COLA, all baseline models are trained solely on  target city bus timetables.
  For COLA, the target city model is trained using only target city bus timetables, while source-city data are varied across bus timetables, mobility trajectories, and bus timetables + mobility trajectories configurations.
  We report COLA's best result across these source-city data configurations.
  In addition, since no explicit road network information is considered,  models that use road topology~\cite{zhu2024controltraj,tao2024map2traj,wei2024diff,cao2025holistic,wei2025transfertraj} are excluded from comparison.
  For all methods, starting locations and times of generated trajectories are randomly sampled from the spatial and temporal distributions of target city bus timetables. 
  All generated trajectories are evaluated on the target city mobility trajectory test sets.
  
\begin{itemize}
    \item \textbf{TimeGeo}~\cite{jiang2016timegeo}: 
  TimeGeo is a classical statistical trajectory generation method that models individual mobility by capturing periodic patterns and regularities in spatial stay and travel times.
  \item \textbf{MoveSim}~\cite{feng2020learning}: 
    MoveSim simulates individual mobility behaviors by learning transition patterns and temporal distributions from real trajectory data.
    \item \textbf{TSG}~\cite{wang2021large}: 
  TSG is a two-stage trajectory generation framework. 
  The first stage learns inter-region transition relationships, while the second stage refines trajectories at the path level using road network information. 
  Since no road network data are used in our setting, we only adopt the first stage to generate  trajectories.
    \item \textbf{TrajGDM}~\cite{chu2023trajgdm}: 
  TrajGDM is a diffusion-based trajectory generation method that learns the spatiotemporal distribution of real trajectories through a progressive denoising process.
    \item \textbf{DiffTraj}~\cite{zhu2023difftraj}: 
  DiffTraj adopts a diffusion-based generative method for modeling human mobility trajectories.
  \item \textbf{TrajGEN}~\cite{cao2021generating}: 
  TrajGEN  focuses on learning the statistical characteristics of trajectories in terms of spatial distributions and temporal rhythms.
  \item \textbf{Traveller}~\cite{luo2025traveller}: 
  Traveller generates individual mobility trajectories via sequence modeling and preserves global structural consistency of trajectories.
  \item \textbf{Geollama}$^{+}$~\cite{li2025geo}: Geo-Llama leverages the strong semantic understanding and reasoning LLMs by reformulating trajectory generation as a sequence prediction problem.
  Unlike the original Geo-Llama design, we treat the ID of each location as a newly introduced spatial token and explicitly train its corresponding embedding and LM head. In addition, the up, down, and gate linear layers in the FFN module are jointly fine-tuned using LoRA. To distinguish it from the original Geo-Llama, we denote this variant with a superscript “+”.
  \item \textbf{COLA}~\cite{wang2024cola}: 
  COLA is a transferable trajectory generation framework that explicitly models cross-city differences. 
  In our setting, the target city is trained using only its bus timetables, while the source city are trained with multiple data combinations, including  bus timetables, mobility trajectories, and bus timetables + mobility trajectories. 
  We report the best-performing configuration, with detailed results for different combinations of WX$\rightarrow$SH shown in Table~\ref{table:cola_results}.
\end{itemize}
\begin{table*}[ht]
  \centering
  \caption{Performance comparison of COLA under different settings.}
  \label{table:cola_results}

    \begin{tabular}{lcccc}
      \toprule
      Method (Setting) & Dist. & Rad. & Dur. & Loc. \\
      \midrule
      COLA (WX-Bus \& WX-Mob , SH-Bus) & 0.1911 & 0.1690 & 0.0316 & 0.1689 \\
      COLA (WX-Bus, SH-Bus)       & 0.1831 & 0.1564 & 0.0344 & 0.1544 \\
      COLA (WX-Mob, SH-Bus)       & 0.1843 & 0.1652 & 0.0335 & 0.1668 \\
      \bottomrule
    \end{tabular}

  \vskip -0.1in
\end{table*}

\section{Evaluation Metrics}
\label{metric detail}
We evaluate model performance by comparing generated trajectories with real mobility trajectories in terms of discrepancies across several  statistical distributions.
Following  prior works, we compute the following statistical properties from both generated and real trajectories:
\begin{itemize}
    \item Distance (Dist.) measures the cumulative geographic distance between consecutive visited locations within a single trajectory.
    \item Radius (Rad.) describes the spatial extent of a trajectory, reflecting the dispersion of an individual’s activity area.
    \item Duration (Dur.) represents the distribution of consecutive dwell times at the same location.
    \item TrajLoc (Loc.) denotes the number of distinct locations visited within a single trajectory and serves as a measure of spatial diversity.
\end{itemize}
For each of the above statistical attributes, we adopt the Jensen–Shannon Divergence (JSD) to quantify the discrepancy between the generated distribution $D'$ and the real distribution $D$.  
JSD is defined as
\begin{equation}
    \mathrm{JSD}(D, D') =
\frac{1}{2}\mathrm{KL}\!\left(D \,\middle\|\, \frac{D + D'}{2}\right)
+
\frac{1}{2}\mathrm{KL}\!\left(D' \,\middle\|\, \frac{D + D'}{2}\right),
\end{equation}
where $\mathrm{KL}(\cdot)$ denotes the Kullback–Leibler divergence.
As all these attributes are continuous-valued, we first estimate their probability distributions using histogram-based density estimation and then compute the corresponding JSD values. 
Smaller JSD values indicate that the generated trajectories more closely match the real trajectories, and therefore exhibit higher realism.

\section{Details of Trajectory Generation Behavior under Different Spatial Data Coverage Conditions}
\label{detail: data coverage}
\begin{figure}
  \centering
  \centerline{\includegraphics[scale=0.25]{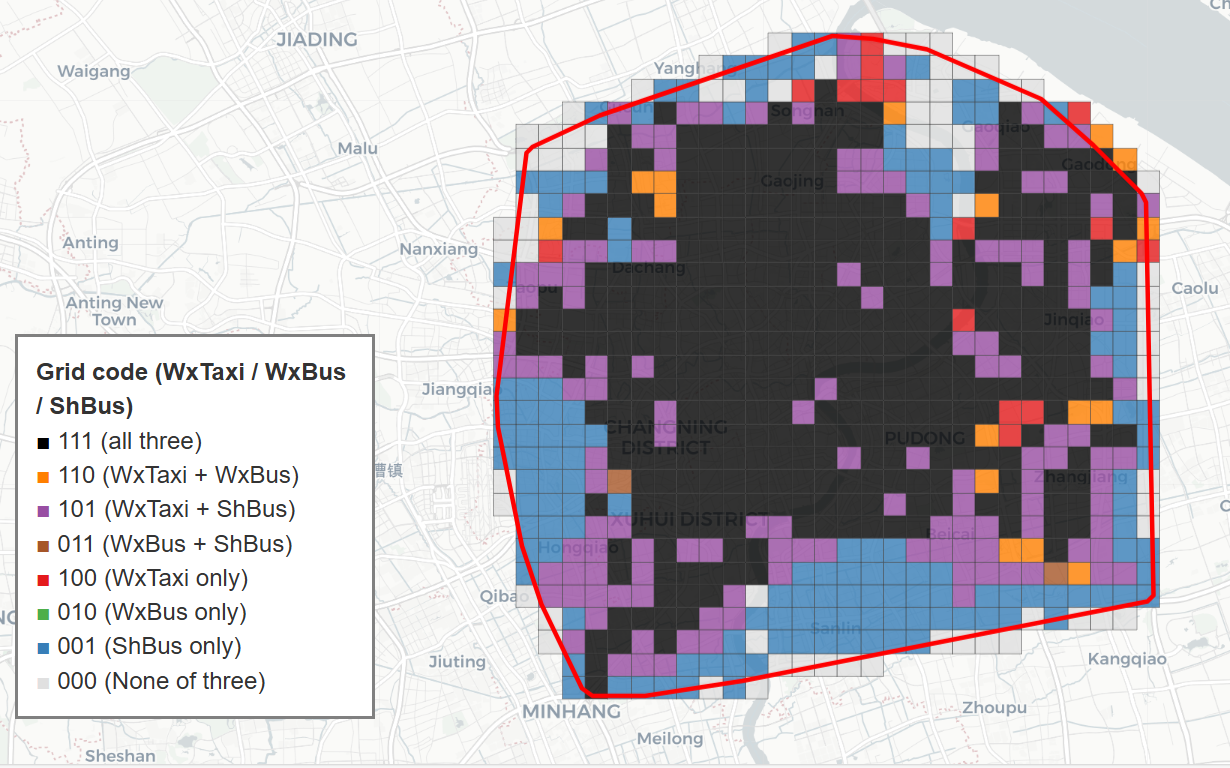}}
    \caption{
        Different categories of Shanghai grids.
    }
    \label{fig:shanghai grids}
\end{figure}

We categorize Shanghai locations based on whether they are covered by WuXi mobility trajectories, Wuxi  bus timetables, and Shanghai bus timetables.
Each location is represented by a three-bit binary code $abc \in \{0,1\}^3$, where $a,b$ and $c$ indicate whether the location is covered in these three source of data.
Fig.~\ref{fig:shanghai grids} illustrates the spatial distribution of these spatial data coverage
categories in Shanghai.

To characterize trajectory generation behavior under these spatial data coverage conditions, we consider three complementary metrics.

We first examine the distribution of stay durations  to assess how well MobTA captures local activity intensity in different spatial data coverage categories. Specifically, each trajectory is segmented into consecutive stay events, and each event is assigned to the corresponding spatial data coverage category based on its location. 
For each category, we compute the duration distributions for both real and generated trajectories and measure their discrepancy using the Jensen–Shannon Divergence (JSD), denoted as  \textbf{JSD$_{dur}$}.

Next, we quantify spatial reachability at the trajectory level by computing the trajectory coverage ratio. 
For each generated or real trajectory, we check whether it visits at least one location grid belonging to a given spatial data coverage category. 
Let $n_{'abc'}^{\text{traj}}$ denote the number of trajectories satisfying this condition in spatial data coverage category $'abc'$ , and let $N$ be the total number of trajectories. 
The trajectory coverage ratio for spatial data coverage category $'abc'$ is defined as
\begin{equation}
    \rho^{\text{traj}}('abc')=\frac{n_{'abc'}^{\text{traj}}}{N}.
\end{equation}

Finally, we analyze visit frequency by measuring the proportion of total visits falling into each spatial data coverage category.
Let $n_{'abc'}^{\text{traj}}$ denote the total number of trajectory visits occurring in spatial data coverage category $'abc'$, and let $N_v$ be the total number of visits across all trajectories. 
The visit frequency is defined as
\begin{equation}
    \rho^{\text{visit}}('abc')=\frac{n_{'abc'}^{\text{visit}}}{N_v}.
\end{equation}
Together, these metrics provide a fine-grained characterization of MobTA’s trajectory generation behavior under different spatial data coverage conditions.

\section{Ablation Study}
\label{ablation}

\subsection{Ablation Study for Sub Task Vectors}
Table~\ref{tab:ablation_elh} presents ablation results of task vectors applied to different parameter subspaces, where E, L, and H indicate whether task vectors are introduced in the embedding layer, the Transformer backbone, and the LM head, respectively.
The results show that the Transformer backbone (L) constitutes the most critical component of the task vector.
Introducing task vectors only in this subspace already leads to substantial improvements.
However, relying solely on L is insufficient to achieve optimal performance.
 Further incorporating task vectors in E and H consistently yields additional performance gains.
These results show that though the dominant contribution of  task vector arises from the Transformer backbone, excluding task vectors in either the embedding layer or the LM head prevents MobTA from achieving optimal generation performance.

\begin{table}[t]
\centering
\caption{Ablation study on different task vector components.}
\label{tab:ablation_elh}

\begin{tabular}{ccc|cccc}
\hline
E & L & H & Dist. & Rad. & Dur. & Loc. \\
\hline
$\times$ & $\times$ & $\times$ & 0.2058 & 0.1493 & 0.0342 & 0.1175 \\
$\times$ & $\times$ & $\checkmark$ & 0.2149 & 0.1557 & 0.0312 & 0.1167 \\
$\times$ & $\checkmark$ & $\times$ & 0.1217 & 0.0979 & 0.0486 & 0.0721 \\
$\times$ & $\checkmark$ & $\checkmark$ & 0.1256 & 0.0971 & 0.0464 & 0.0736 \\
$\checkmark$ & $\times$ & $\times$ & 0.2032 & 0.1400 & 0.0369 & 0.1051 \\
$\checkmark$ & $\times$ & $\checkmark$ & 0.2173 & 0.1509 & 0.0324 & 0.1098 \\
$\checkmark$ & $\checkmark$ & $\times$ & 0.1199 & 0.0929 & 0.0433 & 0.0779 \\
\hline
$\checkmark$ & $\checkmark$ & $\checkmark$ & \textbf{0.0969} & \textbf{0.0639} & \textbf{0.0206} & \textbf{0.0487} \\
\hline
\end{tabular}

\end{table}

\subsection{Necessity of Bus-timetable–guided Mobility Backbone}
This ablation study investigates whether bus-timetable-based trajectory generation is essential for providing a city-level mobility backbone.
We construct a diagnostic setting in which GeoLLaMA$^+$ is trained solely on source-city mobility trajectories (Shanghai or Wuxi) and then directly applied to generate target city (Singapore) mobility trajectories.
As shown in Table~\ref{tab:source_only_taxi}, models trained only on source-city mobility trajectory data can merely capture certain coarse-grained mobility characteristics.
Moreover, SH (mob) $\rightarrow$SG consistently outperforms WX(mob)$\rightarrow$SG across all metrics, indicating that mobility patterns learned from a larger and more structurally complex city exhibit stronger cross-city generalization. 
However, compared with methods that incorporate bus timetables (e.g., COLA and MobTA), this setting exhibits clear degradation in spatial related metrics (Dist., Rad., and Loc.).
These results highlight the critical role of bus timetables in cross-city trajectory generation. 
As a planning-driven and highly stable component of urban transportation systems, bus timetables encode the dominant, city-level mobility backbone. 
In the absence of such data, models trained solely on source-city mobility data struggle to adapt to the spatial layout  of the target city, leading to degraded generation quality. 
\begin{table}[t]
  \centering
  \caption{Transfer results when only source city real mobility trajectories are used for training}
  \label{tab:source_only_taxi}
  \resizebox{\columnwidth}{!}{
  \begin{tabular}{l|cccc}
    \toprule
    Setting & Dist. & Rad. & Dur. & Loc. \\
    \midrule
    SH (mob) $\rightarrow$SG & 0.2241 & 0.2017 & 0.0479 & 0.1466 \\
    WX(mob) $\rightarrow$SG & 0.2749 & 0.2361 & 0.0717 & 0.1998 \\
    \bottomrule
  \end{tabular}}
\end{table}

\section{Hyperparameter Sensitivity Analysis}
\label{hyper}
\begin{figure}
  \centering
  \includegraphics[scale=0.4]{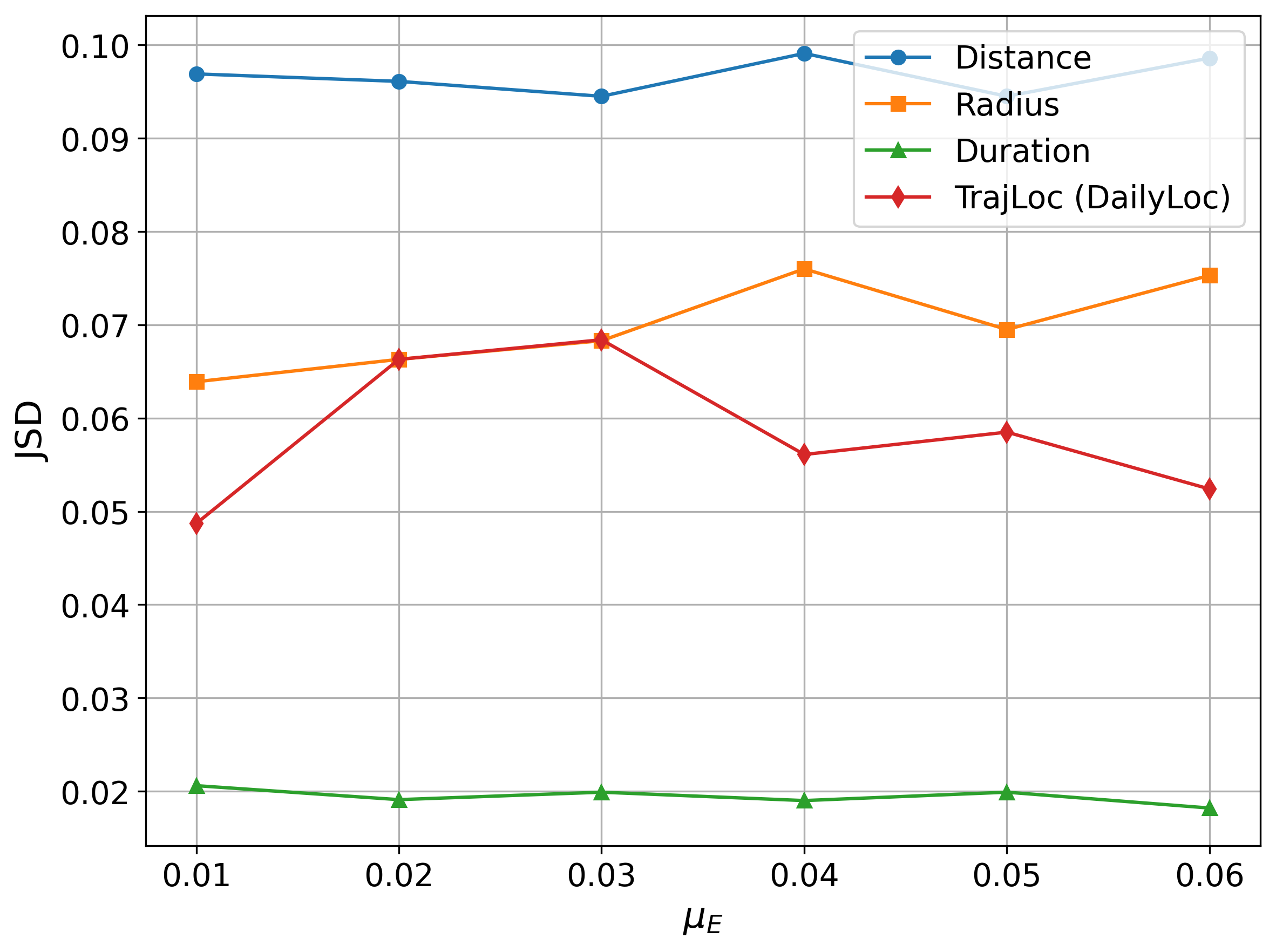}
    \caption{
        Hyperparameter sensitivity for $\mu_E$.
    }
    \label{mu_E_ablation}
\end{figure}

\begin{figure}
  \centering
  \includegraphics[scale=0.4]{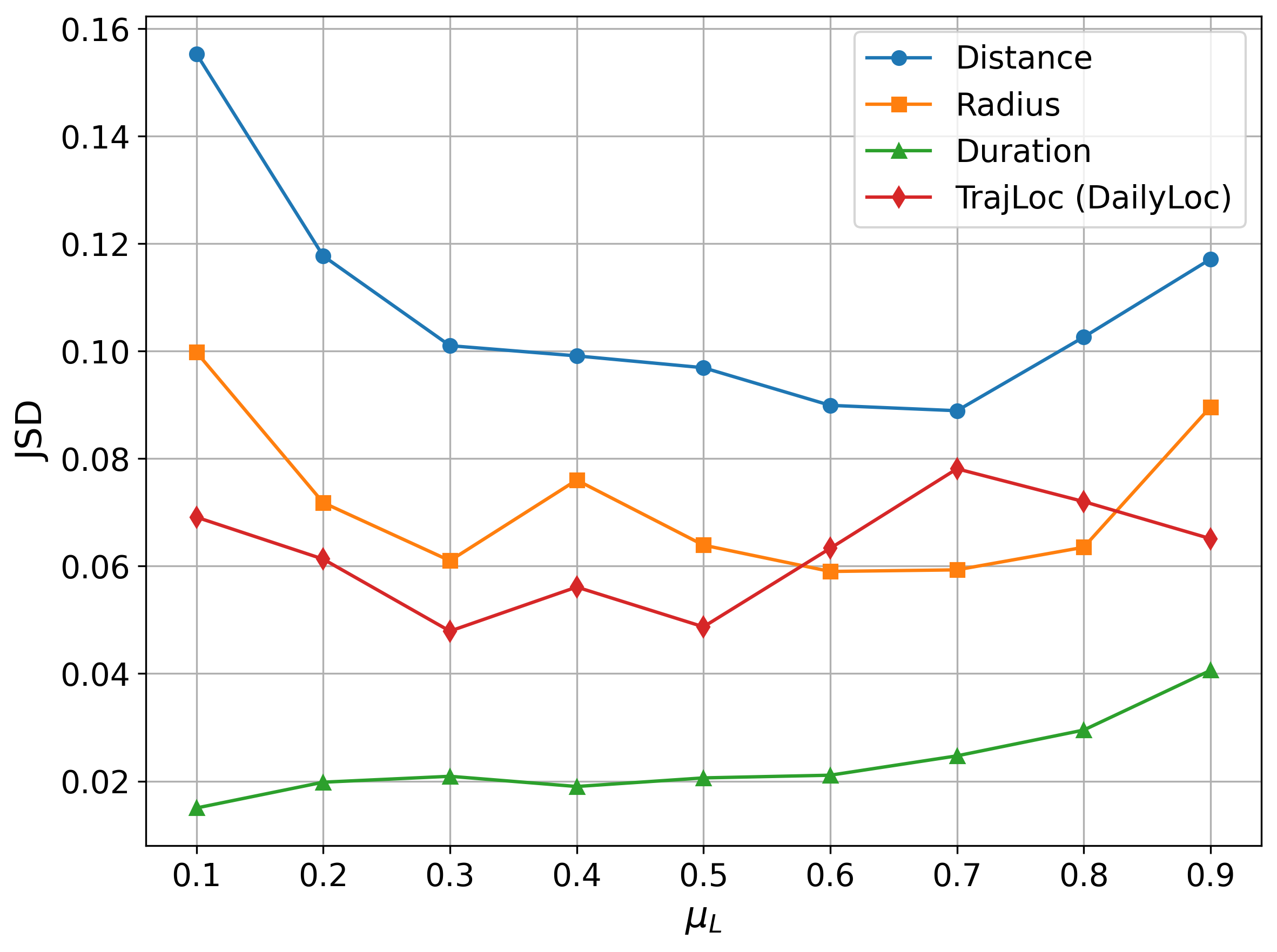}
    \caption{
        Hyperparameter sensitivity for $\mu_L$.
    }
    \label{mu_L_ablation}

\end{figure}

\begin{figure}
  \centering
  \includegraphics[scale=0.4]{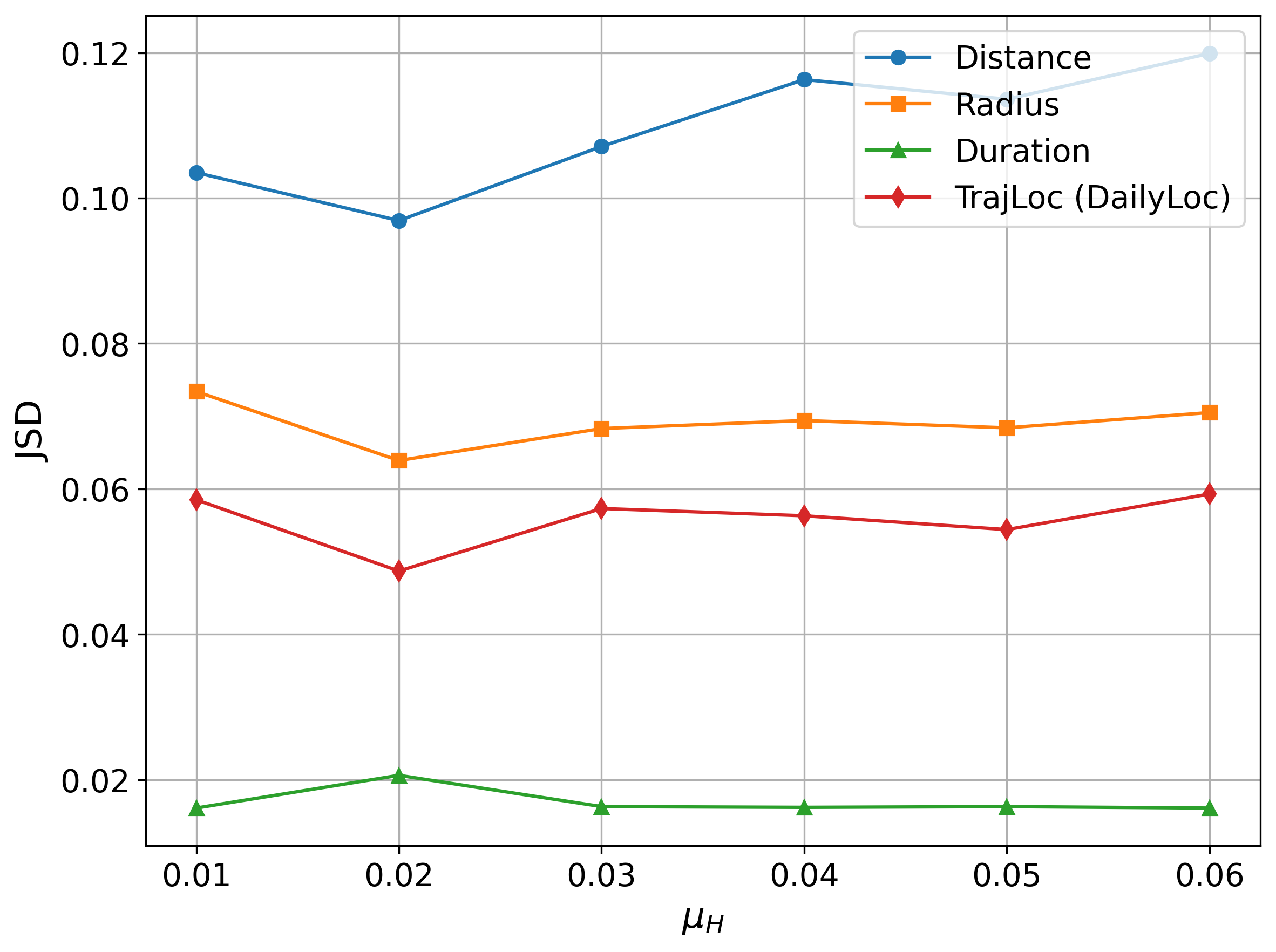}
    \caption{
        Hyperparameter sensitivity for $\mu_H$.
    }
    \label{mu_H_ablation}
\end{figure}

Fig.\ref{mu_E_ablation},\ref{mu_L_ablation},\ref{mu_H_ablation} present the sensitivity analysis of the weighting coefficients $\mu_E,\mu_L,\mu_H$, which control the contributions of task vectors in the embedding layer, the Transformer backbone, and the output layer, respectively.
Overall, MobTA exhibits stable performance across a range of parameter values, indicating that it is not sensitive to hyperparameter choices and demonstrates  robustness.
Among the three coefficients, $\mu_L$ has the most pronounced impact on generation performance. When $\mu_L$ is set too small or too large, the performance degrades noticeably, suggesting that the Transformer backbone plays a central role. In contrast, $\mu_E$ and $\mu_H$ mainly serve as auxiliary adjustment factors, and variations within reasonable ranges have relatively minor effects on overall generation quality.

Based on this sensitivity analysis, we provide a practical guideline for hyperparameter selection in zero-shot scenarios.
We observe that MobTA exhibits performance stability across a broad range of  backbone coefficients $(\mu_L \in [0.2,0.8])$. 
This insensitivity to precise hyperparameter tuning is critical for zero-shot scenarios: it implies that a fixed default configuration (e.g., $\mu_L=0.5$, $\mu_E=0.01, \mu_H=0.02$) can be reliably applied to unseen target cities without the need for validation data. 
This discrepancy in magnitude of $\mu_L$ and $\mu_E,\mu_H$ is because the LLM backbone is updated via LoRA, while the embedding and output head undergo full-parameter fine-tuning.
Consequently, the task vectors derived from full-parameter updates possess larger magnitudes and thus necessitate smaller scaling factors to prevent disrupting the pre-trained feature space.

\section{Implementation Considerations and Technical Nuances}
\label{consider}
To enhance transparency and facilitate reproducibility, we additionally report implementation considerations and technical nuances.

\subsection{Experimental Configurations and Reproducibility Details}
We provide the implementation details and experimental configurations here.
We utilize the Unsloth library to accelerate the fine-tuning of LLM models.
To reduce memory footprint while maintaining optimization stability, we employ the 8-bit AdamW optimizer.
Hyperparameter configurations are shown in Table~\ref{tab:hyper config}:

\begin{table}[t]
\centering
\caption{Hyperparameter Configuration.}
\label{tab:hyper config}
\begin{tabular}{c|c}
\hline
LoRA Rank & 32 \\
LoRA Alpha & 64 \\
LoRA Target Module & Q,K,V,Up,Down,Gate \\
Optimizer & AdamW (8-bit)\\
Learning Rate & 5e-5\\
Batch Size & 4 \\
Num Epochs & 3\\
Max Sequence Length &2048\\
$\mu_E$ & 0.01 \\
$\mu_L$ & 0.5 \\
$\mu_H$ & 0.02 \\
\hline
\end{tabular}

\end{table}

\subsection{Practical Implementation for Querying Bus Stop Locations via  Amap API}
When querying geographic coordinates of urban bus stops using the Amap (Gaode Maps) API, we observe that whether the query keyword explicitly includes the term “bus stop” has a decisive impact on localization accuracy. 
If only the stop name is used as the query keyword, the API frequently returns locations associated with nearby roads, intersections, or surrounding landmarks, with spatial deviations that can exceed several kilometers in certain areas. 

\begin{figure}
  \vskip 0.2in
  \begin{center}
    \centerline{\includegraphics[scale=0.35]{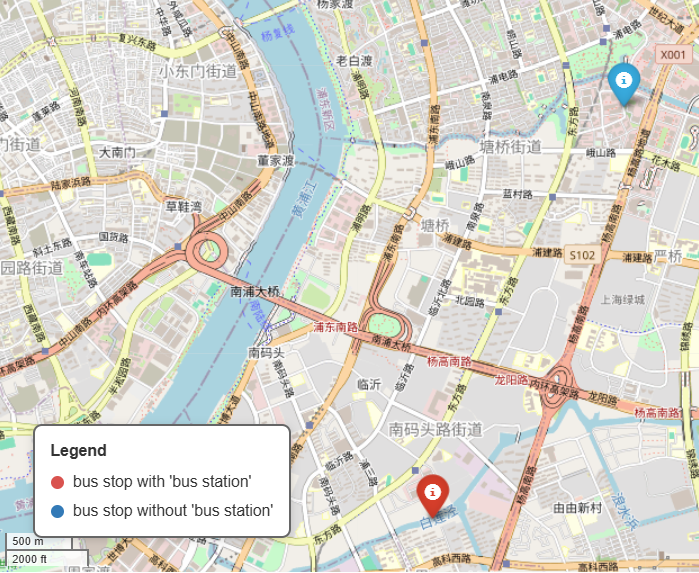}}
    \caption{
        Failure case of Amap API querying.
    }
    \label{gaode fail}
  \end{center}
\end{figure}

In contrast, explicitly appending the keyword “bus stop” to the query consistently yields coordinates that align well with the actual physical locations of bus stops.
Fig.~\ref{gaode fail} illustrates this phenomenon by comparing the returned
coordinates for the Shanghai bus stop “Dongfang Road–Pusan Road” with and
without the “bus stop” keyword, revealing substantial spatial discrepancies.
This issue is particularly pronounced in dense urban areas.
If left unaddressed, it would introduce systematic errors into subsequent spatial discretization and eventually, the performance of trajectory generation.

\subsection{Reproducibility Risks of Unsloth’s New Token “Interpolation” Initialization}
\label{sec:repro}
During implementation, we observed that Unsloth’s function add\_new\_tokens(..., method="interpolation") internally first calls HuggingFace’s tokenizer.add\_tokens(new\_tokens) and model.resize\_token\_embeddings(len(tokenizer)), thereby expanding the vocabulary and initializing the corresponding embedding and LM head parameters.
Only after this step does it enter the so-called “interpolation” initialization logic.

Conceptually, interpolation initialization is intended to first decompose the string form of a new token into a sequence of existing token IDs using the original tokenizer, compute the mean of their embedding vectors to obtain $\bar{e}_{base}$, and then linearly interpolate between this value and the global embedding mean $\bar{e}$:
\begin{equation}
    e_{new}=(1-\alpha)\bar{e}+\alpha\bar{e}_{base}. 
\end{equation}
However, because resize\_token\_embeddings has already been executed prior to interpolation, the tokenizer at this stage already recognizes the new token. 
As a result, a string that should have been decomposed into a sequence of old tokens is instead matched directly to the new token itself. 
Consequently, the embedding vector retrieved during the interpolation step corresponds to the newly and randomly initialized token embedding rather than a semantic average of existing tokens. 
In practice, the resulting initialization is therefore closer to random initialization with linear mixing than to the intended semantic interpolation initialization.

This implementation order also introduces additional reproducibility risks. HuggingFace’s resize\_token\_embeddings randomly initializes embeddings (and corresponding LM head parameters) for newly added tokens. 
If the random seed is not explicitly fixed, even two runs with identical data and configurations may yield different initial embeddings for new tokens. 
These differences can propagate to the embedding and LM head parameters after training and further affect downstream task vectors, ultimately undermining the stability of task vector arithmetic.

Consequently, to avoid such randomness and ensure reproducibility, in MobTA, we
initialize the embeddings and LM head parameters of newly introduced spatial
tokens using the deterministic global mean of the pre-trained LLM’s existing token embeddings and
output head weights, following the implementation provided by Unsloth’s \texttt{mean} initialization option.

\subsection{Automatic Untying of Embedding and LM Head After Adding New Tokens in Unsloth}
\label{sec:automa}
In most mainstream LLM implementations, the input embedding matrix and the output LM head are typically weight-tied, meaning that they share the same parameter matrix to ensure consistency between input and output spaces. 
However, when extending the vocabulary using add\_new\_tokens within the Unsloth framework and subsequently fine-tuning the model with PEFT-based methods, we observe that Unsloth automatically unties the embedding and LM head parameters.

As a result, for newly introduced tokens, the input embedding vectors and the corresponding output-layer weights are treated as two independent trainable parameter subspaces. 
These parameters must be updated separately during training and handled independently during model saving and loading. 
They can no longer be assumed to represent a single shared parameter matrix or be copied interchangeably. 
This behavior is particularly important to account for when analyzing parameter updates and constructing task vectors involving newly added tokens.

\end{document}